\documentclass{article}

\usepackage[margin=3cm]{geometry}

\usepackage{authblk}

\usepackage{microtype}
\usepackage{todonotes}
\usepackage{graphicx}
\usepackage{subfigure}
\usepackage{tikz}
\usetikzlibrary{shapes,arrows}
\usetikzlibrary{matrix}
\usepackage{pgfplots}
\pgfplotsset{compat=newest}
\usepgfplotslibrary{groupplots}
\usepgfplotslibrary{dateplot}
\usetikzlibrary{math}
\tikzmath
{
  function symlog(\x,\a){
    \yLarge = ((\x>\a) - (\x<-\a)) * (ln(max(abs(\x/\a),1)) + 1);
    \ySmall = (\x >= -\a) * (\x <= \a) * \x / \a ;
    return \yLarge + \ySmall ;
  };
  function symexp(\y,\a){
    \xLarge = ((\y>1) - (\y<-1)) * \a * exp(abs(\y) - 1) ;
    \xSmall = (\y>=-1) * (\y<=1) * \a * \y ;
    return \xLarge + \xSmall ;
  };
}
\tikzset{
  block/.style    = {draw, thick, rectangle, minimum height = 3em, minimum width = 3em},
  causalvar/.style      = {draw, circle, node distance = 2cm}
}

\usepackage{booktabs} 
\usepackage{hyperref}

\usepackage{natbib}

\newcommand{\theHalgorithm}{\arabic{algorithm}}
\newcommand{\indep}{\perp \!\!\! \perp}
\newcommand{\ie}{\textit{i.e.} }
\newcommand{\cf}{\textit{c.f.} }
\newcommand*\circled[1]{\tikz[baseline=(char.base)]{
            \node[shape=circle,draw,inner sep=2pt] (char) {#1};}}
            
\newtheorem{defin}{Definition}
\newtheorem{lem}{Lemma}
\newtheorem{prop}{Proposition}

\newtheorem{remark}{Remark}

\usepackage{amsmath}
\usepackage{centernot}
\usepackage{amssymb}
\newcommand{\ind}{\perp\mkern-9.5mu\perp}
\newcommand{\notind}{\centernot{\ind}}
\newcommand{\Egiv}[2]{\mathbb{E}[ #1\given #2]}
\newcommand{\E}[1]{\mathbb{E}[ #1]}
\newcommand{\Edocgiv}[4]{\mathbb{E}^{#1;do(#2)}[#3\given#4]}
\newcommand{\Ecgiv}[3]{\mathbb{E}^{#1}[#2\given#3]}
\newcommand{\Probgiv}[2]{\mathbb{P}(#1\given#2)}
\newcommand{\Prob}[1]{\mathbb{P}(#1)}
\newcommand{\SCM}{\mathfrak{C}}
\newcommand{\Probdoc}[3]{\mathbb{P}^{#1;do(#2)}(#3)}
\newcommand{\Probdocgiv}[4]{\mathbb{P}^{#1;do(#2)}(#3\given#4)}
\newcommand{\Probc}[2]{\mathbb{P}^{#1}(#2)}
\newcommand{\Probcgiv}[3]{\mathbb{P}^{#1}(#2\given#3)}
\newcommand{\user}{x}
\newcommand{\User}{X}
\newcommand{\Treat}{P}
\newcommand{\Outcome}{Y}
\newcommand{\Display}{T}
\newcommand{\Conf}{U}
\newcommand{\Bid}{B}
\newcommand{\given}{|}
\newcommand{\bidf}{\alpha}
\newcommand{\pderiv}[1]{\frac{\partial #1}{\partial \bidf}}
\newcommand{\ra}{\rightarrow}
\newcommand{\la}{\leftarrow}
\newcommand{\no}[1]{\overline{#1}}
\newcommand{\wrt}{\textit{w.r.t.} }
\newcommand{\er}{\widehat{er}}
\newcommand{\pe}{\widehat{pe}}
\newcommand{\Userset}{\mathcal{\User}}
\newcommand{\Dataset}{\mathcal{D}}
\newcommand{\tmreg}{$\mathrm{\Teffect}_{\mathrm{2M}}$}
\newcommand{\tmmed}{$\mathrm{\OurEstimator}_{\mathrm{2M}}$}
\newcommand{\sdrreg}{$\mathrm{\Teffect}_{\mathrm{SDR}}$}
\newcommand{\sdrmed}{$\mathrm{\OurEstimator}_{\mathrm{SDR}}$}
\newcommand{\bestreg}{$\mathrm{\Teffect}_{\mathrm{best}}$}
\newcommand{\sumionen}{\sum\limits_{i=1}^{n}}
\newcommand{\sumi}{\sum_i}
\newcommand{\sumj}{\sum_j}
\newcommand{\TE}[1]{\tau^{#1}}
\newcommand{\TEE}[1]{\hat{\tau}^{#1}}
\newcommand{\compl}{\gamma}
\newcommand{\complE}{\hat{\gamma}}
\newcommand{\interf}{\eta}
\newcommand{\Teffect}{IPE}
\newcommand{\Meffect}{ITE}
\newcommand{\OurEstimator}{C-IPE}

\newcommand{\TR}[1]{\todo[color=Crimson!20!LightGray,author=\textbf{\small TR}]{\small #1\\}}

\title{Individual Treatment Prescription Effect Estimation\\ in a  Low Compliance Setting}
\date{}

\author{Thibaud Rahier, Am\'elie H\'eliou, Matthieu Martin, \\Christophe Renaudin and Eustache Diemert}
\affil{Criteo AI Lab}

\begin{document}
\maketitle

\begin{abstract}
Individual Treatment Effect (\Meffect) estimation is an extensively researched problem, with applications in various domains.
We model the case where there exists heterogeneous non-compliance to a randomly assigned treatment, a typical situation in health (because of non-compliance to prescription) or digital advertising (because of competition and ad blockers for instance).
The lower the compliance, the more the effect of treatment prescription $-$ or individual prescription effect (\Teffect) $-$ signal fades away and becomes hard to estimate.
We propose a new approach for the estimation of the \Teffect~that takes advantage of observed compliance information to prevent signal fading. Using the Structural Causal Model framework and \textit{do-calculus}, we define a general mediated causal effect setting and propose a corresponding estimator which consistently recovers the \Teffect~with asymptotic variance guarantees. Finally, we conduct experiments on both synthetic and real-world datasets that highlight the benefit of the approach, which consistently improves state-of-the-art in low compliance settings. 
\end{abstract}

\section{Introduction}
\label{introduction}

Individual Treatment Effect (ITE) estimation is an important task in various applications such as healthcare \citep{foster2011subgroup}, online advertising \citep{diemert2018large} or socio-economics  \citep{xie2012estimating}. As it is often the case in practice, we assume that we cannot directly control the treatment intake but only the treatment prescription: we therefore focus on the Individual Prescription Effect (\Teffect), which designates the effect of the \textit{treatment prescription} $\Treat$ on the \textit{outcome} $\Outcome$ for an individual described by \textit{covariates} $\User$ (\cf Equation~\eqref{eq:ITE}, assuming random prescription assignment).
\begin{equation}
\label{eq:ITE}
\begin{split}
    \Teffect(\user) =& ~\E{\Outcome|\User=\user,\Treat=1} - \E{\Outcome|\User=\user,\Treat=0}
\end{split}
\end{equation}
We also assume that we observe the \textit{treatment intake} (or equivalently, the compliance to prescription), denoted by $\Display$, which acts as a \textit{mediator} of the causal effect of $\Treat$ on $\Outcome$. 
The effect of $\Display$ on $\Outcome$ will henceforth be referred to as \Meffect.

Table~\ref{table:intro} illustrates various settings in which non-compliance to treatment prescription may occur \citep{gordon2019comparison, jin2008factors, parker2000reducing}. This happens for instance when individuals have the choice not to abide by the prescription or if there exists conflicting interests. Of course one can choose to focus the study on actually treated individuals only (\Meffect). But from a decision making point of view it often makes sense to consider that future treatment decisions need to take into account the possibility of non-compliance so as to accurately predict future expected outcomes. For example a policy maker would want to take into account that not all individuals would abide by the new policy (as can be estimated from a pilot study) to predict the expected impact of a roll-out of said policy.

\begin{table}[t!]
\label{table:intro}
\caption{Examples of Covariates ($\User$), Outcome ($\Outcome$), Treatment Prescription ($\Treat$)
and Evidence of Treatment Acceptance ($\Display$)}
\label{sample-table}
\begin{center}
\begin{scriptsize}
\begin{sc}
\begin{tabular}{lccc}
\toprule
Var. & Medicine & Online adv. & Job training  \\
\midrule
$\User$     &  patient info       & purchase history     & schooling         \\
$\Treat$     & drug prescription     & bid placement     & training offer        \\
$\Display$     &  drug intake          & ad displayed      & training done    \\
$\Outcome$     & recovery              & sale/visit        & employment \\
\end{tabular}
\end{sc}
\end{scriptsize}
\end{center}
\end{table}

Now we argue that \Teffect~estimation can be hampered by non-compliance. Indeed, note that individuals who did not actually receive the treatment contribute only through noise to the default \Teffect~estimator, as their observed outcome is not effectively influenced.
The variance of the \Teffect~estimator therefore increases as the compliance level decreases. 

Besides, individual treatment (whether prescription or intake) effect models are often considered as prescriptive tools. Indeed, treatment effect predictions can be used in order to target treatment to individuals for which it is the most beneficial \citep{devriendt2018literature, radcliffe2011real}. This calls for an evaluation metric that measures by how much the average treatment effect would have been improved, had the treatment been targeted not by a random instrument, but by the predictions of the considered model instead. For that purpose \cite{rzepakowski2012decision}, \cite{rzepakowski2010decision}, and \cite{radcliffe2011real} have proposed the \textit{Area Under the Uplift Curve} (AUUC) metric that sums the benefits over individuals ranked by predictions. An interesting property of this metric is that it can be used on real data for which we observe a given individual in either treated or untreated conditions but never both.

Confronted with the challenges of (i) learning  \Teffect~models in conditions of (possibly high) non-compliance and (ii) evaluating them as prescriptive tools, we pose the problem in the setting of causal inference and derive an \Teffect~estimator that takes advantage of observed compliance. Our main contributions are as follows.
\begin{enumerate}
    \item Formalization of \Teffect~estimation in a setting of observed compliance described using structural causal models, which extends to more general cases of mediated causal effect estimation (Section \ref{sec:problem_statement})
    \item Proposition of a meta-estimator (in which can be plugged any \Meffect~estimator) for \Teffect~estimation in presence of non-compliance, proof of consistency (Section~\ref{ssec:M-IPEE_def}) and asymptotic variance properties (Section \ref{ssec:asympt_props})
    \item Thorough empirical evaluation of this meta-estimator on synthetic and real world datasets (Section~\ref{sec:expe})
\end{enumerate}

\section{Related work}\label{sec:related_works}

We review three main domains that are concerned with research questions similar to our work: ITE modeling, non compliance in causal inference and evaluation metrics for ITE modeling.

First, we note that ITE models are a pervasive concept in different research fields such as marketing - under the name uplift models \citep{radcliffe2011real}, statistics - as conditional average treatment effect estimators \citep{kunzel2019metalearners} or econometrics  - heterogeneous treatment effect models \citep{jacob2019group, wager2018estimation}.
A simple yet highly scalable approach consists in learning a regression of $\Outcome$ on $\User$ separately in both treatment ($\Treat=1$) and control ($\Treat=0$) populations and return the difference, known as T-learner studied by \cite{kunzel2019metalearners} or ``Two Models'' studied by \cite{radcliffe2011real}. A variation of this approach with larger model capacity have been proposed through a shared representation (SDR) for the treatment and control group by \cite{betlei2018uplift}.
Also, a prolific series of work exists on adapting decision trees and random forests to the causal inference framework \citep{athey2016recursive, wager2018estimation, athey2019generalized}. Further in the same vein and building on work done on double machine learning by \cite{chernozhukov2018double}, \cite{oprescu2018orthogonal} generalize the idea of causal forests, allowing for high-dimensional confounding. 
Finally, another recent trend is to study theoretical limits in ITE estimation and especially generalization bounds \citep{shalit2017estimating, alaa2018limits}.

Then regarding the concept of non compliance, many algorithms have been studied in order to recover (individual) causal effects in non-compliance settings, however 
to our knowledge, works tackling this problem (\textit{e.g.} \cite{gordon2019comparison}) focus solely on effect of the treatment \textit{intake} $\Display$ $-$ and not the treatment \textit{prescription} $\Treat$ $-$ on the outcome $\Outcome$. In that context, the effect of $\Treat$ on $\Outcome$ is sometimes studied in an \textit{instrumental variable} framework to recover the effect of $\Display$ on $\Outcome$ \citep{angrist1995identification, syrgkanis2019machine}. Our work however focuses on estimating the effect of the treatment prescription $\Treat$ on the variable $\Outcome$, taking advantage of the observed treatment intake $\Display$, in the spirit of individual \textit{intention-to-treat} (ITT) analysis \citep{montori2001intention, gupta2011intention} or per-protocol effect \citep{hernan2017per}.
The idea of taking advantage of a mediation variable for causal effect estimation has triggered many recent works \citep{hill2015measuring}, in settings sometimes similar to ours \citep{loeys2015cautionary}. However, the associated assumptions ($\Display$ and $\Outcome$ are unconfounded) are more restrictive than the ones we propose, as we allow all the covariates ($\User$) to be confounders between the mediation variable $\Display$ (representing the treatment intake) and $\Outcome$, and state a result holding for individual (and not only average) treatment effect.

Finally, treatment effect estimators performance is typically done using synthetic data, in which a pointwise error measure $-$ Precision Estimation of Heterogeneous Effect (PEHE) defined by \cite{shalit2017estimating} $-$ is available. However in real world cases, the \textit{fundamental problem of causal inference} states that the ground truth of individual treatment effect cannot be observed (since an individual is either treated or untreated but never both at the same time), preventing to use such metrics beyond simulation studies.
Since our main motivation is to determine which individuals are good candidates for treatment \textit{prescription}, we choose to evaluate the performance of our estimators on real data using the AUUC metric, which evaluates the ranking of individuals implied by corresponding \Meffect~or \Teffect~predictions. One can view the resulting measure as a prediction of the expected benefit of prescribing treatment according to the model prediction instead of a random uniform prescription. Overall, AUUC has been used in recent years in machine learning research to evaluate baseline ITE models vs SDR \citep{diemert2018large}, flavors of Support Vector Machines for ITE estimation \citep{kuusisto2014support} or direct treatment policy optimization \citep{yamane2018uplift}. A normalized variant of AUUC 
also exists under the name Qini coefficient introduced by \cite{radcliffe2011real}, but we choose to focus on AUUC in this work.

\section{Framework}\label{sec:problem_statement}

We briefly recall causality notions used throughout the paper, such as \textit{structural causal model}, \textit{causal graph}, \textit{intervention} and \textit{valid adjustment set} first introduced by \cite{pearl2009causality}, and more recently presented by \cite{peters2017elements}.

\begin{defin}{Structural Causal Model (SCM)} A SCM of variables $\mathbf{X} = \{X_1, \dots, X_d\}$ is an object $\SCM := (\mathbb{S}, \mathbb{P}_\mathbf{N})$ where:\\
(1) $\mathbb{S}$ is a set of $d$ structural assignments $X_i = f_i(\mathbf{PA}_i, N_i)$, with the $f_i$'s deterministic functions and $\mathbf{PA}_i$ the set of parents (direct causes) of $X_i$.\\
(2) $\mathbb{P}_{\mathbf{N}} = \mathbb{P}_{N_1,\ldots,N_d}$ is a joint distribution  over the noise variables $\{N_i\}_{1\leq d}$, which we require to be jointly independent.
\end{defin}
As shown in Chapter~6 of \cite{peters2017elements}, a SCM $\SCM$ induces a unique \emph{causal graph} $\mathcal{G}_{\SCM}$ $-$ defined as the directed acyclic graph (DAG) obtained by creating a vertex for each $X_i$ and drawing directed edges from each $\mathbf{PA}_i$ to $X_i$ (thus justifying the term `parents' for the sets $\mathbf{PA}_i$) $-$ and a unique \emph{entailed distribution} $\mathbb{P}^{\SCM}$ over variables $X_1, \dots, X_d$ such that for each $i$, $X_i = f_i(\mathbf{PA}_i, N_i)$ in distribution.

A SCM can be used to define \emph{interventional distributions}. A (hard) intervention $do(X_i = x_i)$ is a forced assignment of variable $X_i$ to the value $x_i$, which implies a change in the distribution of the variables $X_1, \dots, X_d$. Formally, an intervention $do(X_i = x_i)$ is equivalent to modifying $\SCM$ in two ways: (1) change the structural assignment $X_i = f_i(X_i, \mathbf{PA}_i)$ to $X_i = x_i$ in $\mathbb{S}$ and (2) replace $X_i$ by $x_i$ in all other structural assignments implying $X_i$ in $\mathbb{S}$.\\
We will denote by $\mathbb{P}^{\SCM;do(X_i:=x_i)}$ the probability distribution entailed by the SCM $\SCM$ under the intervention $do(X_i:=x_i)$.

Let $X_i, X_j \in \mathbf{X}$, $\mathbf{Z} \subseteq \mathbf{X} \backslash \{X_i, X_j\}$ is called a \emph{valid adjustment set} (VAS) for the ordered pair $(X_i, X_j)$, if
$$
\Probdoc{\SCM}{X_{i}}{X_{j}} = \sum\limits_{\mathbf{z}}\Probcgiv{\SCM}{X_{j}}{X_{i}, \mathbf{z}}\Probc{\SCM}{\mathbf{z}},
$$
where the sum is over the range of $\mathbf{Z}$.

\noindent\textbf{Notations}. For sake of compactness, we use the following notations for any binary variable $W$ and multi-dimensional variable $X$: $\Prob{W} \triangleq \Prob{W = 1}$, $\Prob{\no{W}} \triangleq \Prob{W = 0}$,  $\Prob{x} \triangleq \Prob{X = x}$, $\text{do}(W) \triangleq \text{do}(W:=1)$.

\subsection{Non-compliance setting}\label{ssec:SCM_def}
The setting of non-compliance we consider in this work is entirely defined by a SCM of variables $\User, \Treat, \Display, \Outcome, \Conf$ for which example values where proposed in Table~\ref{table:intro}:
$\User$, belonging to a multi-dimensional space $\Userset$, contains the individual's descriptive features, or \textit{covariates} (by simplicity, we will confuse individuals and their covariates, referring for example to `an individual $\user$'), 
$\Treat$ is the binary treatment \textit{prescription} variable, 
$\Outcome$ is the binary \textit{outcome} variable,
$\Display$ is the binary treatment \textit{intake} (or acceptation) $-$ which acts as a \textit{mediator} of the causal effect of $\Treat$ on $\Outcome$ $-$
and
$\Conf$ represents (allowed) \textit{unobserved confounders} between $\User$ and $\Outcome$.

\begin{figure}[ht]
\vspace{.25in}
\begin{center}
\begin{tikzpicture}[auto, thick, node distance=1cm, >=triangle 45]
\tikzstyle{unobserved}=[thick, dashed, fill=gray!20]
\tikzstyle{norn}=[thick, fill=gray!20]
\draw
	node[norn] at (0,0)[causalvar](Z){$\Treat$}
	node[norn] at (2,0)[causalvar](M){$\Display$}
	node[norn] at (4,0)[causalvar](Y){$\Outcome$}
	node[norn] at (2,1.5)[causalvar](X){$\User$}
	node[unobserved] at (3.5,1.5)[causalvar](U){$\Conf$};
	\draw[->](Z) -- node {} (M);
	\draw[->](M) -- node {} (Y);
	\draw[->](X) -- node {} (M);
	\draw[->](X) -- node {} (Y);
	\draw[->](U) -- node {} (X);
	\draw[->](U) -- node {} (Y);
\end{tikzpicture}
\caption{\textbf{Causal Graph $\mathcal{G}_\SCM$ Induced by SCM $\SCM$}}
\label{figure:SCM}
\end{center}
\vspace{.25in}
\end{figure}
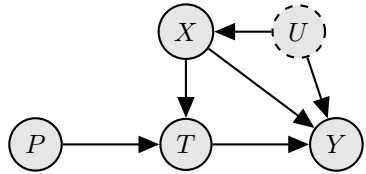

In what follows, we define the structural causal model $\SCM = (\mathbb{S},\mathbb{P}_\mathbf{N})$, which is henceforth assumed to represent the causal mechanisms underlying the variables of interest in this work. The associated causal graph $\mathcal{G}_\SCM$ is given in Figure~\ref{figure:SCM}.

$\mathbb{S}$ is defined in Equations~\eqref{eq:SCMdef}:
\begin{align}\label{eq:SCMdef}
\begin{split}
    \Treat &= \tilde{N}_\Treat \\
    \Conf &= N_\Conf\\
    \User &= f_\User(\Conf, N_\User) \\
    \Display &= f_\Display(\User, N_\Display) \times \Treat\\
    \Outcome &= f_\Outcome(\User, \Display, \Conf, N_\Outcome).
\end{split}
\end{align}
$\mathbb{P}_\mathbf{N}$ satisfies the following mild conditions: $N_\Conf, N_\User, N_\Display, N_\Outcome$ are noise consistent with variables definitions, and $\tilde{N}_\Treat$ is distributed according to a Bernoulli distribution with parameter $p = \Probc{\SCM}{\Treat}$, consistent with a \textit{randomized controlled experiment} setting.

In the next proposition, we list four assumptions implied by $\SCM$ about the variables of interest.
\begin{prop}\label{prop:assumptions}
The SCM $\SCM$ defined in Equations~\eqref{eq:SCMdef} implies the following assumptions on variables $\Treat$, $\Display$, $\Outcome$ and $\User$:
\begin{align}\label{eq:assumptions}
\begin{split}
\textit{(Randomized prescription) }&\ \  \Treat \indep \User\\
\textit{(Exclusive treatment effect) }&\ \ \Treat \indep \Outcome \ \given \ \{\User, \Display\}\\
\textit{(One-sided non-compliance) }&\ \ \Treat = 0 \Rightarrow \Display = 0 \\
\textit{(Valid covariate adjustment) }&\ \ \{\User\} \text{ is a VAS for }(\Display,\Outcome)    
\end{split}
\end{align}
\end{prop}

The complete proof of Proposition~\ref{prop:assumptions}, including definitions of the associated notions, is given in appendix.

\paragraph{Detailed review of assumptions mentioned in Proposition~\ref{prop:assumptions}}. We now review in details the assumptions presented in Equation~\eqref{eq:assumptions} and their connections to applications and usual theoretical settings.
\begin{itemize}
    \item \emph{Randomized prescription} translates the fact that each individual is randomly allocated to a prescription ($\Treat = 1$) or a non-prescription/control ($\Treat = 0$) group, allowing us to infer the causal effect of $\Treat$ on $\Outcome$ without worrying about confounding bias.  This is a typical situation in experiments such as A/B tests or randomized controlled clinical trials.
    \item \emph{Exclusive treatment effect} captures the fact that the prescription to treatment $\Treat$ has an effect on the outcome $\Outcome$ only if the treatment is actually taken $\Display=1$. More generally, this assumption translates the fact that the effect of $\Treat$ on $\Outcome$ is fully mediated by $\Display$. This is typically assumed to be the case in online advertising, in which case $\Treat$ is the allocation to a given advertising engine, and $\Display$ is the actual exposure to advertisement from this engine, as in the study of \cite{gordon2019comparison}.
    \item \emph{One-sided non-compliance} defined by \cite{gordon2019comparison} states that $\Treat =0 \implies \Display = 0$, but that $\Treat = 1 \centernot\implies \Display = 1$. In other words, treatment is only taken ($\Display = 1$) in cases where it was originally assigned ($\Treat = 1$). This assumption is typically satisfied in the instrumental variable setting where $\Treat$ plays the role of the instrument for the estimation of the effect of $\Display$ on $\Outcome$ as explained by \cite{syrgkanis2019machine}.
    \item \emph{Valid covariate adjustment}: every confounder for the effect of $\Display$ on $\Outcome$ is observed (and contained in $\User$). This assumption is classically made in causal inference works \citep{alaa2018limits, alaa2019validating} and is sometimes referred to as \emph{ignorability} \citep{shalit2017estimating}. Under the potential outcome framework \citep{rubin2005causal}, this assumption is equivalent to 
$$
\Outcome(\Display=1), \Outcome(\Display=0) \indep \Display \given \User.
$$
Strong ignorability (where we additionally assume that $0 < \Probgiv{\Display}{\user} <1$ for all $\user$) has been proven to be sufficient to recover the individual causal effect of $\Display$ on $\Outcome$ \citep{imbens2009recent, pearl2017detecting}.
\end{itemize}

\paragraph{Relation to the IV setting}. 
The \emph{One-sided non-compliance} assumption \citep{gordon2019comparison} $-$ stating that $\Treat =0 \implies \Display = 0$, but that $\Treat = 1 \centernot\implies \Display = 1$, \ie treatment is only taken if it was originally prescribed $-$ is typically satisfied in the \textit{instrumental variable} (IV) setting, where $\Treat$ plays the role of the instrument for the estimation of the effect of $\Display$ on $\Outcome$ \citep{syrgkanis2019machine}. There are essential differences with our work, notably (i) we are interested in the effect of $\Treat$ on $\Outcome$ (\Teffect) and not in the effect of $\Display$ on $\Outcome$, and (ii) \textit{exclusive treatment effect} $-$ stating that the effect of $\Treat$ on $\Outcome$ is fully mediated by $\Display$ $-$ is not assumed in typical IV settings.

\subsection{\Teffect~ in presence of non-compliance}\label{ssec:ITE_func}
\noindent\textbf{Notations}.
For all $\user \in \Userset$, we formally define the individual prescription effect $\TE{\Teffect}(\user)$, the individual treatment effect $\TE{\Meffect}(\user)$ and the \textit{individual compliance} probability $\compl(\user)$ as follows:
\begin{align}\label{eq:notations}
\begin{split}
\TE{\Teffect}(\user) &= \Probdocgiv{\SCM}{\Treat}{\Outcome}{\user} - \Probdocgiv{\SCM}{\no{\Treat}}{\Outcome}{\user},\\
\TE{\Meffect }(\user) &= \Probdocgiv{\SCM}{\Display}{\Outcome}{\user} - \Probdocgiv{\SCM}{\no{\Display}}{\Outcome}{\user},\\
\compl(\user) &= \Probdocgiv{\SCM}{\Treat}{\Display}{\user}.    
\end{split}
\end{align}
We also define the relative \Meffect~$\beta(\user)$ and relative \Teffect~$\alpha(\user)$ as:
\begin{align*}
    \alpha(\user) &= \left(\Probcgiv{\SCM}{\Outcome}{\Treat, \user} - \Probcgiv{\SCM}{\Outcome}{\no{\Treat}, \user}\right) \big/ \ \Probcgiv{\SCM}{\Outcome}{\no{\Treat}, \user}, \\
    \beta(\user) &= \left(\Probcgiv{\SCM}{\Outcome}{\Display, \user} - \Probcgiv{\SCM}{\Outcome}{\no{\Display}, \user} \right) \big/ \ \Probcgiv{\SCM}{\Outcome}{\no{\Display}, \user}.
\end{align*}
We propose to exploit the mediation variable $\Display$ by splitting the $\Treat$ to $\Outcome$ path into two \emph{subpaths} (from $\Treat$ to $\Display$ and from $\Display$ to $\Outcome)$, both with a higher signal-to-noise ratio. Indeed, under $\SCM$ we can insert $\Display$ into the $\Probdocgiv{\SCM}{\Treat}{\Outcome}{\user}$ expression as presented in the next lemma.
\begin{lem}\label{lemma1}
Assuming $\SCM$, and for any $\user \in \Userset$, the positive outcome probability under treatment, $\Probdocgiv{\SCM}{\Treat}{\Outcome}{\user}$, can be written as follows:
\begin{equation}
    \label{decompConversionRate}
    \Probdocgiv{\SCM}{\Treat}{\Outcome}{\user} =  \Probcgiv{\SCM}{\Outcome}{\user, \no{\Display}}
    \ + \ \TE{\Meffect }(\user)\compl(\user). 
\end{equation}
\end{lem}
The proof of Lemma~\ref{lemma1} is fully detailed in appendix. It relies on \emph{valid covariate adjustment}, \emph{randomized prescription}, and \emph{exclusive mediation} assumptions, that we have proven to be implied by $\SCM$ in Proposition~\ref{prop:assumptions}.

The first term on the right hand side of Equation~\eqref{decompConversionRate} 
is the individual probability of positive outcome of an individual $\user$ given they did not receive treatment, and is referred to as the \emph{organic} probability of positive outcome.\\
In a nutshell: for a given individual, Lemma~\ref{lemma1} states that their probability of positive outcome given treatment prescription is equal to their organic probability of positive outcome, plus the product of their individual treatment (intake) effect and their compliance.

In Proposition~\ref{prop:decompITE}, we present a result linking the \Teffect, the \Meffect~and the individual compliance:
\begin{prop}\label{prop:decompITE}
Assuming $\SCM$ and for any $\user \in \Userset$, the \Teffect~decomposes as follows:
\begin{equation}
    \label{decompUpliftWithDos}
\TE{\Teffect}(\user) = \TE{\Meffect }(\user)\compl(\user)\\
\end{equation}
\end{prop}
The proof of Proposition~\ref{prop:decompITE} is fully detailed in appendix. It relies on the identity $\Probdocgiv{\SCM}{\no{\Treat}}{\Outcome}{\user} = \Probcgiv{\SCM}{\Outcome}{\user, \no{\Display}}$, which holds under $\SCM$ thanks to the \emph{exclusive treatment effect} and \emph{one-sided non-compliance} assumptions.\\
In intuitive terms, Proposition~\ref{prop:decompITE} states that the effect of treatment prescription on a given individual is equal to the effect of treatment intake on this individual, multiplied by their compliance.

\section{Proposed Approach}\label{sec:novel_estimator}

\subsection{Definition and basic properties}\label{ssec:M-IPEE_def}

The expression proven in Proposition \ref{prop:decompITE} calls for a novel way to estimate $\TE{\Teffect}(\user)$, by first estimating separately both factors $\TE{\Meffect}(\user)$ and $\compl(\user)$, then multiplying these estimators to form a \emph{compliance-aware individual prescription effect} (\OurEstimator) estimator.\\
Formally, let $\TEE{\Meffect}$ be an estimator of $\TE{\Meffect}$ and $\complE$ be an estimator of $\compl$. We define the associated \OurEstimator~estimator $\TEE{\OurEstimator}$, for any $\user$, as:
\begin{equation}\label{M-IPE}
    \TEE{\OurEstimator}(\user) = \TEE{\Meffect }(\user)\complE(\user).
\end{equation}

In practice, $\TEE{\Meffect}$ may be any individual treatment effect estimator. Indeed, under $\SCM$, the individual causal effect of $\Display$ on $\Outcome$ given $\User$ is identifiable since $\{\User\}$ is a valid adjustment set for $(\Display, \Outcome)$ as explained in Section~\ref{sec:problem_statement}. \OurEstimator~is not an estimator per say, but rather a \textit{meta-estimator}: a black-box in which any classical \Meffect~and compliance estimators can be plugged-in.

Assuming that for all $\user$, $\TEE{\Meffect}(\user)$ and $\complE(\user)$ are \emph{consistent} estimators of resp. $\TE{\Meffect}(\user)$ and $\compl(\user)$, Proposition~\ref{prop:decompITE} then ensures that $\TEE{\OurEstimator}(\user)$ is a \emph{consistent} estimator of the associated $\TE{\Teffect}(\user)$.

\subsection{Asymptotic variance properties}
\label{ssec:asympt_props}
Thanks to its expression as a function of an \Meffect~estimator, the \OurEstimator~estimator \emph{focuses} on the individuals who actually received treatment, who are exclusively contributing to the 
signal (\emph{Exclusive treatment effect} assumption in Proposition~\ref{prop:assumptions}). We therefore expect the \OurEstimator~estimator to have lower variance than any standard \Teffect~estimator which does not exploit observable compliance.\\
Comparing \OurEstimator~and \Teffect~estimators is all the more fair than we use an analogous version of the standard \Teffect~estimator $\TEE{\Teffect}$ for the \Meffect~estimator $\TEE{\Meffect}$ which is plugged in the \OurEstimator~estimator $\TEE{\OurEstimator}(\user)$. Doing so is feasible since the \emph{valid covariate adjustment} assumption guarantees that both $\TE{\Teffect}$ and $\TE{\Meffect}$ are identifiable and therefore estimable with standard treatment effect estimators. We refer to this approach as \emph{symmetrically formed estimators comparison}, and use it to conduct our experiments in Section~\ref{sec:expe}.\\
In the following proposition, we compare the asymptotic variance of estimators $\TEE{\OurEstimator}$ and $\TEE{\Teffect}$ in the following simple yet realistic setting:

\noindent\textbf{Single-stratum setting}. We focus on the \Teffect~estimation for a single value $\user_0$ of $\User$, for which we assume to observe $n$ \textit{i.i.d.} samples $\{(\user_0, \Treat_i, \Display_i, \Outcome_i)\}_{1\leq i \leq n}$. All stated results generalise to any stratum $S\subset \Userset$ for which there exists $\user_S \in \Userset$ such that the reducted feature variable $\User' \triangleq \user_S \mathbb{I}_{\{\User \in S\}} + \User \mathbb{I}_{\{\User \notin S\}}$ still defines a valid adjustment set for $(\Display,\Outcome)$.


\noindent\textbf{Notations}. Consistently with notations presented in Equations~\eqref{eq:notations}, $\alpha(\user_0)$, $\beta(\user_0)$ refer respectively to the relative \Teffect~and relative \Meffect~in stratum $\{\User = \user_0\}$ (and are assumed to be positive in this illustrative setting), and we denote $\complE(\user_0)$, $\TEE{\Teffect}(\user_0)$ and $\TEE{\Meffect }(\user_0)$ respectively the maximum-likelihood estimator (MLE) of $\compl(\user_0)$, and the MLE-based two-model estimators (difference of two MLE estimators) of $\TE{\Teffect}(\user_0)$ and $\TE{\Meffect }(\user_0)$. We define the associated \OurEstimator~estimator as $\TEE{\OurEstimator}(\user_0) \triangleq \complE(\user_0)\TEE{\Meffect }(\user_0)$. Note that we use the MLE-based Two-Model estimator for both $\TEE{\Meffect}$ and $\TEE{\Teffect}$, thus satisfying the \textit{symmetrically formed estimators comparison} setting.
Lastly, we denote $p_1(\user_0) = \Probcgiv{\SCM}{\Outcome}{\Treat, \user_0}$.\\
In the following Proposition, we present an asymptotic bound for the ratio of the standard deviation ($sd$) of \OurEstimator~and \Teffect~estimators.

\begin{prop}
\label{prop:bound}
Under $\SCM$ defined in Section \ref{ssec:SCM_def} with $\Probc{\SCM}{\Treat} = \frac{1}{2}$, assuming we observe $n$ \textit{i.i.d.} samples in stratum $\{\User = \user_0\}$, we have:
\begin{equation}
\lim_{n\to\infty} \frac{sd(\TEE{\OurEstimator})}{sd(\TEE{\Teffect})} \leq \sqrt{\left(\frac{2(1 + \beta)}{(1 - p_1)(2 + \alpha)}\right)\compl}
\end{equation}
where we dropped the reference to $\user_0$ for clarity.
\end{prop}
This theoretical bound shows that the ratio of standard deviations of \OurEstimator~and \Teffect~estimators gets smaller when the compliance factor $\compl$ decreases, which was expected.

Additional assumptions however need to be made on $\beta$ and $p_1$ in order to derive an informative bound in practice. In real-world datasets (presented in Section~\ref{sec:expe} and in appendix), we consistently observe $\hat{\beta}(\user) \leq 12$ and $\hat{p}_1(\user) \leq 0.05$, with estimators described in Section~\ref{sec:expe}. Under such assumptions on the orders of magnitude of $\beta$ and $p_1$, we can derive a useful-in-practice bound, as presented in the following remark.
\begin{remark}\label{rem:bound}
If we additionally assume $\beta \leq 12$, and $p_1 \leq 0.05$ we have:
\begin{equation}\label{eq:bound_practice}
\lim_{n\to\infty} \frac{sd(\TEE{\OurEstimator})}{sd(\TEE{\Teffect})} \leq 4\sqrt{\compl}.
\end{equation}
where we dropped the reference to $\user_0$ for clarity.
\end{remark}
This bound is derived from loose upper-bounds on $\beta$ and $p_1$ and is only presented for illustrative purposes. It is however still informative in case of low compliance $\compl$. For instance, if $\compl \leq 10^{-2}$, Equation~\eqref{eq:bound_practice} reveals that the asymptotic standard deviation of $\TEE{\OurEstimator}$ is more than twice smaller than the one of $\TEE{\Teffect}$.

\section{Experiments}\label{sec:expe}
To qualify the performance of the \OurEstimator~estimator, we study its benefits in a variety of settings.
Firstly we study its properties on simulation-based studies, hereafter denoted by `Synthetic Datasets', for which (i) the \Teffect~ground truth is known (ii) the level of compliance can be controlled and (iii) we can appreciate performance with respect to an Oracle.
Lastly we apply our approach to transform baseline \Teffect~estimators and compare their performance (AUUC) on three real-world, large scale datasets: an open dataset from Criteo, named `CRITEO-UPLIFT1 Dataset'\footnote{\scriptsize{http://cail.criteo.com/criteo-uplift-prediction-dataset/}}, and two \emph{private} datasets on which we have privileged access, designated by `Private datasets' and for which full results are given in appendix.

In each experiment we take care of comparing symmetrically formed estimators: we consider baseline treatment effect estimators which we use both as (i) an \Teffect~estimator and as (ii) an \Meffect~estimator to be plugged in the \OurEstimator~decomposition. To simplify experiments we chose two baseline estimators: Two Models (2M) and Shared Data Representation (SDR) as they easily scale to large datasets and have been found competitive in prior studies by \cite{betlei2018uplift}. For reproducibility sake we have implemented all models using the Scikit-Learn Python library \citep{pedregosa2011scikit} and provide all code as supplementary material. All experiments were run on a machine with 48 CPUs (Intel(R) Xeon(R) Gold 6146 CPU @ 3.20GHz), with 2 Threads per core, and 500Go of RAM.
Finally, we note that the state of the art is always evolving and improving. We did not use the most advanced models because we do not aim at outperforming them. Instead, we claim that the \OurEstimator~estimator can improve any \Teffect~estimator if plugged-in as the \Meffect~factor in \eqref{M-IPE} in a low compliance setting.

\subsection{Datasets}
\paragraph{Synthetic Datasets}
\label{ssec:synthexpe}
We define a simulation setting in which $\Userset = \{0,1\}^{10}$, $n=2.10^6$. The outcome is generated according to:
\begin{equation}
Y~\sim~\mathrm{Bern}\left( p_0 \left(1 + \Treat \Display \beta(x)\right) \right),
\label{synt_outcome}
\end{equation}
where $\Treat~\sim~\mathrm{Bern}(0.5)$, $\Display~\sim~\mathrm{Bern}(\compl(\user))$, and $p_0 = \Probcgiv{\SCM}{\Outcome}{\no{\Display}, \user} = 0.1$, using notations from Equations~\eqref{eq:notations}. This procedure allows for varying $\compl(\user)$ and $\beta(\user)$ to simulate different levels of compliance and relative \Meffect, respectively.

\begin{figure}[!ht]
\vspace{.25in}
\begin{center}
\centerline{\includegraphics[width=.60\columnwidth]{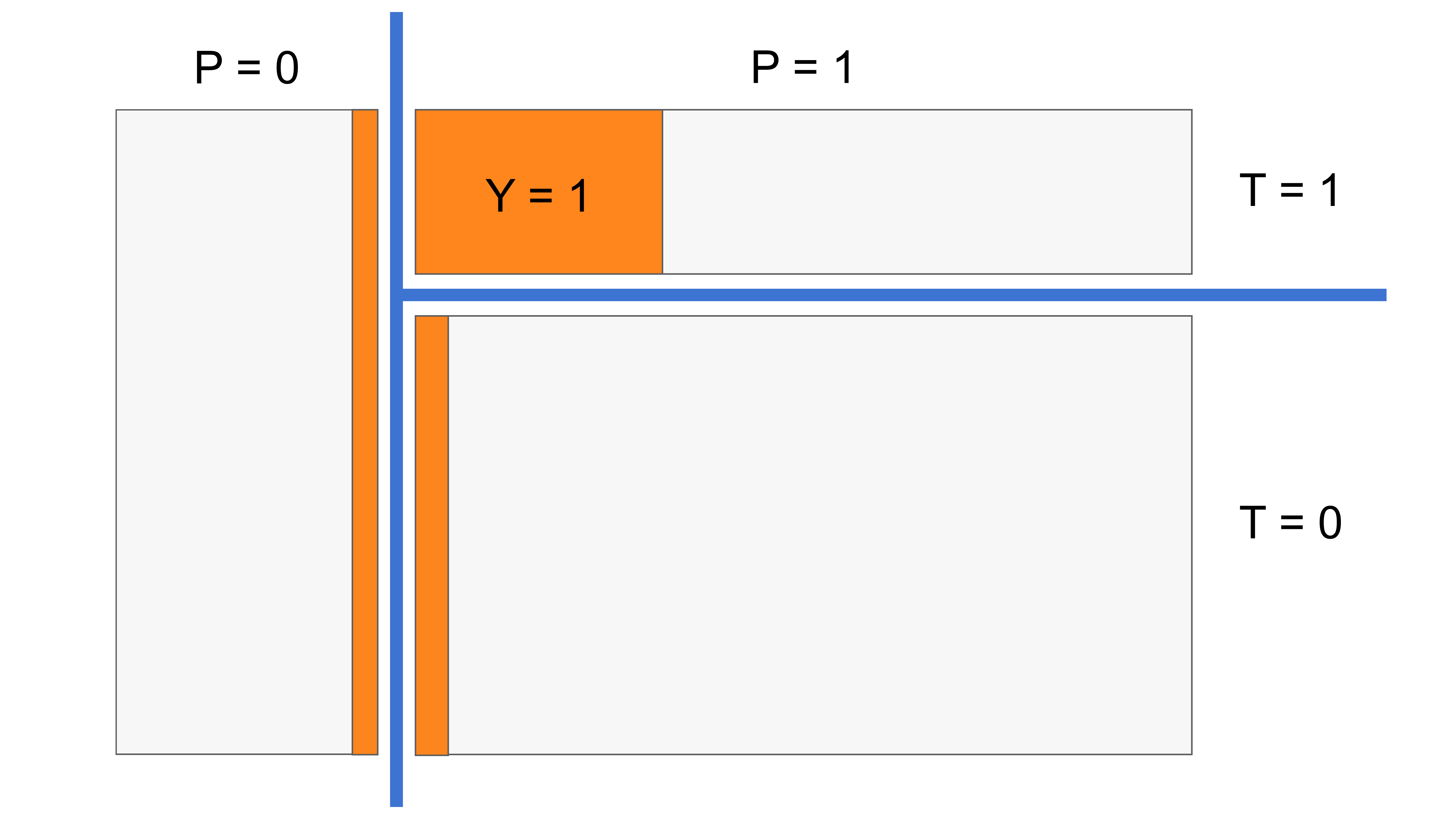}}
\vspace{.25in}
\caption{Illustration of the \textbf{Exclusive Treatment Effect Assumption} in the CRITEO-UPLIFT1 Dataset: the effect of $\Treat$ on $\Outcome$ goes mainly through the realization of $\Display$}
\label{figure:signal-to-noise}
\end{center}
\end{figure}

\paragraph{CRITEO-UPLIFT1 Dataset}
\label{ssec:realworldexpe}
This open dataset from Criteo contains online advertising data from a randomized controlled trial. Key statistics for this dataset are summarized in Table~\ref{tab:perTreatment} and~\ref{tab:perExpo}. Notably, average treatment prescription $\E\Treat \approx .85$ indicates that $15\%$ of users were assigned to the control group and shown no advertisement. For the rest of the population, the advertisers participated in online ad auctions. Among the users that advertisers tried to expose ($\Treat=1$), only  $3.65\%$ actually saw an ad, which corresponds to a very low average compliance level $\E{\compl(\User)}$, which we expect to highlight \OurEstimator~estimator benefits. Effective exposure to ads is encoded by the $\Display$ variable in this setup, while the outcome~$\Outcome$ encodes `at least one visit on the advertiser website'. As illustrated in Figure~\ref{figure:signal-to-noise}, the mean of $\Outcome$ is more than $10$ times higher for exposed users ($\Display=1$) than for non-exposed ones ($\Display=0$).

\begin{table}[t!]
\caption{Distribution of Users Effectively Treated ($\Display=1$) and Visits ($\Outcome=1$) on CRITEO-UPLIFT1 Split on Treatment Groups}
\label{tab:perTreatment}
\begin{center}
\begin{scriptsize}
\begin{sc}
\begin{tabular}{l|ccc}
\toprule
  treatment ($\Treat$) & exposure ($\Display$) & outcome ($\Outcome$)  \\
\midrule
  0 (2'096'236) & -  & 3.82\% (79'986) \\
  1 (12'161'477) & 3.65\% (444'384) & 4.93\% (599'170) \\
\end{tabular}
\end{sc}
\end{scriptsize}
\end{center}
\end{table}

\begin{table}[t!]
\caption{Distribution of Visits ($\Outcome=1$) on CRITEO-UPLIFT1 Projected on Exposure Groups}
\label{tab:perExpo}
\begin{center}
\begin{sc}
\begin{tabular}{l|ccc}
\toprule
  exposure ($\Display$) & outcome ($\Outcome$)  \\
\midrule
  0 (13'813'329) &  3.58\% (495'003) \\
  1 (444'384) &  \textbf{41.4\%} (184'153) \\
\end{tabular}
\end{sc}
\end{center}
\end{table}

\paragraph{Private Datasets}
We have access to two private datasets of roughly 90M instances, which contain binary treatment prescription $\Treat$ and outcome $\Outcome$, and a suitable variable $\Display$ that embodies an observable compliance. They have a higher average compliance $\compl$ ($7.8\%$ and $10.4\%$), higher number of features (48) and similar signal strength than the CRITEO-UPLIFT1 dataset. The purpose of extending the study to these datasets is to verify the relationship between compliance level and performance, predicted by the theoretical study in Section \ref{ssec:asympt_props}. The corresponding data has been collected in the same geographical location, during two separate time periods).

\subsection{Experiments}

In all experiments, nuisance models required by the 2M and SDR estimators are learned as regularized logistic regressions using second order cross-features. This is also the case for the additional nuisance model needed for the \OurEstimator~estimator: the compliance $\hat{\gamma}(.)$. Hyperparameters (regularization norm and strength) of each of these models is carefully selected using internal cross validation, as detailed in appendix.

\paragraph{Compliance Sensitivity Experiment (Simulation)}
We aim to highlight how the compliance level $\compl$ influences the performance of both standard \Teffect~estimators and \OurEstimator~estimators. For this purpose we use a wide range of values $\compl \in [10^{-4};.99]$, and generate synthetic datasets as described in Section~\ref{ssec:synthexpe} with a different relative \Meffect~values $\beta \in \{-1,0,1,2,3,4,5,6,7,8\}$ for each of the $10$ contexts. We report the PEHE metric for both \tmreg~and \tmmed~estimators, and approximate their variance by repeating the experiment with 51 random test/train splits. Recall that PEHE is the mean squared difference between the \Teffect~ground truth and the estimators predictions.

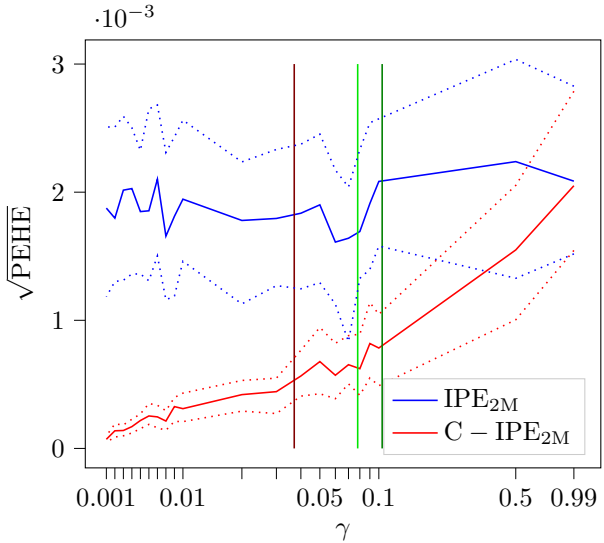
\begin{figure}[!ht]
\vspace{.25in}
\begin{center}
\begin{tikzpicture}
\def\basis{1}
  \pgfplotsset
  {
    x coord trafo/.code={\pgfmathparse{symlog(#1,\basis)}\pgfmathresult},
    x coord inv trafo/.code={\pgfmathparse{symexp(#1,\basis)}\pgfmathresult},
    xticklabel style={/pgf/number format/.cd,int detect,precision=2},
}
\begin{axis}[
legend cell align={left},
legend style={fill opacity=1, draw opacity=1, text opacity=1, at={(0.97,0.03)}, anchor=south east, draw=white!80!black},
tick align=outside,
tick pos=left,
x grid style={white!69.0196078431373!black},
xlabel={$\compl$},
xmin=-0.147514467904021, xmax=135.87614985037,
xtick style={color=black},
xtick={99,50,10,9,8,7,6,5,4,3,2,1,0.9,0.8,0.7,0.6,0.5,0.4,0.3,0.2,0.1},
xticklabels={0.99,0.5,0.1,,,,,0.05,,,,0.01,,,,,,,,,0.001},
y grid style={white!69.0196078431373!black},
ylabel={$\sqrt{\mathrm{PEHE}}$},
ymin=-0.00015182239639494, ymax=0.00318827032429374,
ytick style={color=black}
]

\addplot [semithick, blue]
table {%
99 0.00208586615867097
50 0.00223848699809638
10 0.00208391938366315
9 0.0019112163530788
8 0.0016930164370965
7 0.00164050628579966
6 0.00161048649206431
5 0.00190116790634709
4 0.00183563149785689
3 0.00179494067090752
2 0.00177924476236556
1 0.00194542874708067
0.9 0.00181160071599462
0.8 0.00165679989216312
0.7 0.00210145091273515
0.6 0.00185465424237245
0.5 0.00184845262965951
0.4 0.00202811542567336
0.3 0.00201588440117299
0.2 0.00179735201889245
0.1 0.0018756351821261
};
\addlegendentry{\tmreg}
\addplot [semithick, red]
table {%
99 0.00205028638634087
50 0.00154919163557073
10 0.000783763147404394
9 0.000818855568458748
8 0.000622174544249115
7 0.000653508314774297
6 0.000571163116714479
5 0.000677825284463114
4 0.000565174221340387
3 0.000443160937790108
2 0.000419862989937366
1 0.000310361944729703
0.9 0.000326456591119115
0.8 0.000213719513375667
0.7 0.000246202125502931
0.6 0.000253427170623936
0.5 0.000217775352561916
0.4 0.000170089339087441
0.3 0.000139922602750513
0.2 0.000137617916976365
0.1 7.1226533648842e-05
};
\label{pgfplots:solid}
\addlegendentry{\tmmed}
\addplot [semithick, red!50!black]
table {%
3.7 0.003
3.7 0
};
\label{pgfplots:green}
\addplot [semithick, green!90!black]
table {%
7.8 0.003
7.8 0
};
\label{pgfplots:priv}
\addplot [semithick, green!50!black]
table {%
10.4 0.003
10.4 0
};
\label{pgfplots:priv2}
\addplot [semithick, blue, dotted]
table {%
99 0.00151842898442021
50 0.00132810484972593
10 0.00157962292528341
9 0.00139739102108795
8 0.0013274412053419
7 0.000833328889129729
6 0.00112580570433533
5 0.00129375290147149
4 0.00124595489723232
3 0.00127036930170685
2 0.00112797769301794
1 0.00146094885756006
0.9 0.00118891611260277
0.8 0.00115965068572773
0.7 0.00150279235908915
0.6 0.00130842326754032
0.5 0.00136185988218605
0.4 0.00136128074216764
0.3 0.00131276031719561
0.2 0.00130234833630156
0.1 0.00118401645918582
};
\addplot [semithick, blue, dotted]
table {%
99 0.00282719022855484
50 0.0030364479278988
10 0.00257250024897625
9 0.00254183035777965
8 0.00233249294888509
7 0.00203993818472233
6 0.00217286961621635
5 0.0024532333731729
4 0.00237433559295731
3 0.00233332188484929
2 0.00223612963660535
1 0.00256229583000302
0.9 0.00243091411103783
0.8 0.00231008943713201
0.7 0.00268035863455274
0.6 0.00264977857759062
0.5 0.00232550369583511
0.4 0.00250272427938508
0.3 0.00259225358666449
0.2 0.00250880853413617
0.1 0.00250865923200346
};

\addplot [semithick, red, dotted]
table {%
99 0.00154533681453168
50 0.00100416123355394
10 0.000489488151533601
9 0.000550271764941775
8 0.000415228598601292
7 0.00050424929330989
6 0.000388463356145067
5 0.000426615286203005
4 0.00040833955817941
3 0.000272121727615666
2 0.000291156807315702
1 0.000209111915591712
0.9 0.000211214601499044
0.8 0.00014142113893536
0.7 0.00016114873344905
0.6 0.000187504449124983
0.5 0.000154342377297183
0.4 0.000120786830100671
0.3 9.84382453284501e-05
0.2 8.88193180772388e-05
0.1 4.75622089538504e-05
};
\addplot [semithick, red, dotted]
table {%
99 0.00278745797485485
50 0.00205284102397722
10 0.00104646888411076
9 0.00113884924997145
8 0.000891612362253879
7 0.000880378023844134
6 0.000820416859389436
5 0.000944467604870247
4 0.000766045958901208
3 0.000548443676965184
2 0.000530528130384678
1 0.000431425925610103
0.9 0.000397307140718613
0.8 0.000294019756487329
0.7 0.000336823394038649
0.6 0.000350605302028509
0.5 0.00026895573585356
0.4 0.00022760237192698
0.3 0.000183539881854115
0.2 0.000191083266366252
0.1 9.85465465860842e-05
};
\label{pgfplots:dashed}
\end{axis}

\end{tikzpicture}
\caption{\textbf{Compliance Sensitivity (Simulation study)}. PEHE (lower is better) of \Teffect~vs \OurEstimator~estimators at varying compliance level $\compl$. Solid line \ref{pgfplots:solid} represents the median, and dashed line \ref{pgfplots:dashed} represents 5\% and 95\% of confidence intervals. \ref{pgfplots:green}, \ref{pgfplots:priv} and \ref{pgfplots:priv2}: compliance level $\compl$ for resp. CRITEO-UPLIFT1 and private datasets 1 \& 2. The x-axis has a simlog scale.}
\label{fig:Syn_PEHE}
\end{center}
\vspace{.25in}
\end{figure}

We observe on Figure~\ref{fig:Syn_PEHE} that the \OurEstimator~estimator significantly outperforms the standard \Teffect~estimator when the level of compliance $\compl$ is low, and has similar performance when $\compl$ gets closer to $1$. This shows that our compliance-aware approach significantly reduces the noise induced by non-compliance and is able to detect smaller \Teffect~signals. This is true in particular for compliance levels $\compl$ in the range that is observe on the real datasets that we are using.

\begin{figure}[ht]
\vspace{.25in}
\begin{center}
\pgfplotsset{width=\columnwidth,height=5cm}
\begin{tikzpicture}

\definecolor{color0}{rgb}{1,0.498039215686275,0.0549019607843137}
\definecolor{color1}{rgb}{0.172549019607843,0.627450980392157,0.172549019607843}

\begin{axis}[
tick align=outside,
tick pos=left,
x grid style={white!69.0196078431373!black},
xmin=0.5, xmax=6.5,
xtick style={color=black},
xtick={1,2,3,4,5,6},
xticklabels={\bestreg,\tmreg,\tmmed,\sdrreg,\sdrmed,Oracle},
x tick label style={rotate=90},
y grid style={white!69.0196078431373!black},
ylabel={$\Delta$AUUC},
ymin=3.94885287499996, ymax=419.116850625,
ytick style={color=black}
]
\addplot [black]
table {%
0.75 127.1411625
1.25 127.1411625
1.25 212.001438
0.75 212.001438
0.75 127.1411625
};
\addplot [black]
table {%
1 127.1411625
1 72.130782
};
\addplot [black]
table {%
1 212.001438
1 252.5213405
};
\addplot [black]
table {%
0.875 72.130782
1.125 72.130782
};
\addplot [black]
table {%
0.875 252.5213405
1.125 252.5213405
};
\addplot [black, mark=*, mark size=1, mark options={solid,fill opacity=0}, only marks]
table {%
1 36.556
1 22.8201255
1 49.1027845
1 314.796625
1 308.1454265
1 304.5089415
};
\addplot [black]
table {%
1.75 127.1411625
2.25 127.1411625
2.25 212.001438
1.75 212.001438
1.75 127.1411625
};
\addplot [black]
table {%
2 127.1411625
2 72.130782
};
\addplot [black]
table {%
2 212.001438
2 252.5213405
};
\addplot [black]
table {%
1.875 72.130782
2.125 72.130782
};
\addplot [black]
table {%
1.875 252.5213405
2.125 252.5213405
};
\addplot [black, mark=*, mark size=1, mark options={solid,fill opacity=0}, only marks]
table {%
2 36.556
2 22.8201255
2 49.1027845
2 314.796625
2 308.1454265
2 304.5089415
};
\addplot [black]
table {%
2.75 240.46986875
3.25 240.46986875
3.25 317.69033175
2.75 317.69033175
2.75 240.46986875
};
\addplot [black]
table {%
3 240.46986875
3 191.8415775
};
\addplot [black]
table {%
3 317.69033175
3 361.838461
};
\addplot [black]
table {%
2.875 191.8415775
3.125 191.8415775
};
\addplot [black]
table {%
2.875 361.838461
3.125 361.838461
};
\addplot [black, mark=*, mark size=1, mark options={solid,fill opacity=0}, only marks]
table {%
3 182.3321185
3 144.1746585
3 189.099115
3 361.9974425
3 396.943615
3 387.0884375
};
\addplot [black]
table {%
3.75 127.1411625
4.25 127.1411625
4.25 212.001438
3.75 212.001438
3.75 127.1411625
};
\addplot [black]
table {%
4 127.1411625
4 72.130782
};
\addplot [black]
table {%
4 212.001438
4 252.5213405
};
\addplot [black]
table {%
3.875 72.130782
4.125 72.130782
};
\addplot [black]
table {%
3.875 252.5213405
4.125 252.5213405
};
\addplot [black, mark=*, mark size=1, mark options={solid,fill opacity=0}, only marks]
table {%
4 36.556
4 22.8201255
4 49.1027845
4 314.796625
4 308.1454265
4 304.5089415
};
\addplot [black]
table {%
4.75 240.46986875
5.25 240.46986875
5.25 316.88802825
4.75 316.88802825
4.75 240.46986875
};
\addplot [black]
table {%
5 240.46986875
5 191.8415775
};
\addplot [black]
table {%
5 316.88802825
5 361.838461
};
\addplot [black]
table {%
4.875 191.8415775
5.125 191.8415775
};
\addplot [black]
table {%
4.875 361.838461
5.125 361.838461
};
\addplot [black, mark=*, mark size=1, mark options={solid,fill opacity=0}, only marks]
table {%
5 182.3321185
5 144.1746585
5 189.099115
5 361.9974425
5 396.943615
5 387.0884375
};
\addplot [black]
table {%
5.75 240.46986875
6.25 240.46986875
6.25 317.69033175
5.75 317.69033175
5.75 240.46986875
};
\addplot [black]
table {%
6 240.46986875
6 191.8415775
};
\addplot [black]
table {%
6 317.69033175
6 361.838461
};
\addplot [black]
table {%
5.875 191.8415775
6.125 191.8415775
};
\addplot [black]
table {%
5.875 361.838461
6.125 361.838461
};
\addplot [black, mark=*, mark size=1, mark options={solid,fill opacity=0}, only marks]
table {%
6 183.9474705
6 143.8259705
6 189.099115
6 361.9974425
6 400.245578
6 387.0884375
};
\addplot [color0]
table {%
0.75 165.0213765
1.25 165.0213765
};
\addplot [color1, mark=triangle*, mark size=3, mark options={solid}, only marks]
table {%
1 169.886046686275
};
\addplot [color0]
table {%
1.75 165.0213765
2.25 165.0213765
};
\addplot [color1, mark=triangle*, mark size=3, mark options={solid}, only marks]
table {%
2 169.886046686275
};
\addplot [color0]
table {%
2.75 273.734713
3.25 273.734713
};
\addplot [color1, mark=triangle*, mark size=3, mark options={solid}, only marks]
table {%
3 276.836104882353
};
\addplot [color0]
table {%
3.75 165.0213765
4.25 165.0213765
};
\addplot [color1, mark=triangle*, mark size=3, mark options={solid}, only marks]
table {%
4 169.886046686275
};
\addplot [color0]
table {%
4.75 273.734713
5.25 273.734713
};
\addplot [color1, mark=triangle*, mark size=3, mark options={solid}, only marks]
table {%
5 276.804642
};
\addplot [color0]
table {%
5.75 279.6694865
6.25 279.6694865
};
\addplot [color1, mark=triangle*, mark size=3, mark options={solid}, only marks]
table {%
6 278.716155745098
};
\end{axis}

\end{tikzpicture}
\caption{\textbf{Baseline Experiment (Simulation)}. $\Delta$AUUC (higher is better) of two \Teffect~estimators, corresponding \OurEstimator~estimators and Oracle model (theoretical truth). Box plots are done on 51 random splits, whiskers at 5/95 percentiles. Note how \OurEstimator~systematically increases AUUC of base estimators.}
\label{Syn_AUUC}
\end{center}
\vspace{.25in}
\end{figure}
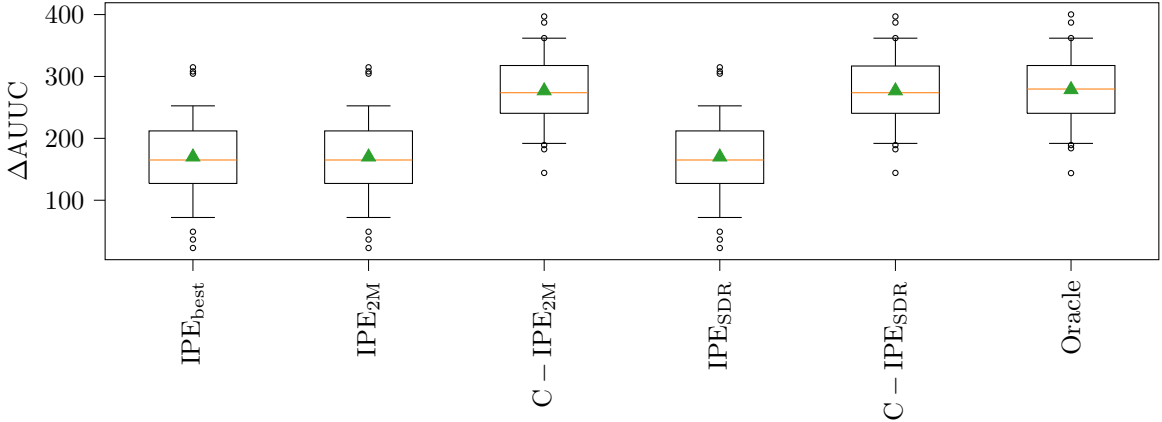

\paragraph{Baseline Experiment (Simulation)}
We simulate a realistic scenario where there exists heterogeneity in compliance $\compl$ and treatment effects $\TE{\Teffect}$. More precisely, for each instance value $\user$, we draw once and for all $\compl \in \{0.01, 0.005\}$ and $\beta \in \{-1, 1, 3, 5, 7\}$ uniformly and independently. The associated outcome $\Outcome$ is computed according to~\eqref{synt_outcome}.
We study four methods: standard two-models (\tmreg), shared data representation (\sdrreg)  and their compliance-aware variants (\tmmed, and \sdrmed).
We focus on the AUUC metric, which measures the capacity of the model to rank individuals according to their \Teffect~. The AUUC has the important advantage of not requiring access to the causal effect ground truth, which is typically the case in real-world applications. 
For more meaningful performance representation, 
we substract the AUUC of a `random model' to all raw AUUC values, obtaining $\Delta$AUUC. Finally, for scaling purposes, we also report in Figure~\ref{Syn_AUUC} results for an Oracle estimator that has access to the drawn $(\beta, \gamma)$, and for \bestreg, the best possible estimator which does not exploit the observable compliance $\Display$ (it predicts for each $\user$ its empirical \Teffect~average based on the training set). Again, variance is estimated by repeating the experiment with 51 random test/train splits.

Figure~\ref{Syn_AUUC} assesses the performance of \Teffect~estimators versus their \OurEstimator~variants, using the $\Delta$AUUC metric. \OurEstimator~estimators yield a higher $\Delta$AUUC on more than 90\% of the random splits.
Moreover the Oracle (best model possible) does not significantly outperform \OurEstimator~estimators, note that even the Oracle can misrank users because the validation set is noisy and empirical \Teffect s do not always follow the expected ranking.
Overall, Figure~\ref{Syn_AUUC} does not show any limitation of the \tmreg~and \sdrreg~estimators, but rather highlights the ineffectiveness of such direct \Teffect~estimators in low compliance settings. This phenomenon can be improved by our compliance-aware approach thanks to the higher signal of the causal effect of $\Display$ on $\Outcome$. Of course, this synthetic data encodes a simpler setting than real-world data, but the fact that our proposed compliance-aware approach performs that well still confirms our theoretical analysis.

\begin{figure}[!h]
\vspace{.15in}
\begin{center}
\pgfplotsset{width=.7\columnwidth,height=5cm}
\begin{tikzpicture}

\definecolor{color0}{rgb}{1,0.498039215686275,0.0549019607843137}
\definecolor{color1}{rgb}{0.172549019607843,0.627450980392157,0.172549019607843}

\begin{axis}[
tick align=outside,
tick pos=left,
x grid style={white!69.0196078431373!black},
xmin=0.5, xmax=4.5,
xtick style={color=black},
xtick={1,2,3,4},
xticklabels={\tmreg,\tmmed,\sdrreg,\sdrmed},
y grid style={white!69.0196078431373!black},
ylabel={\(\displaystyle \Delta\)AUUC},
ymin=20619.2194386436, ymax=22874.1032621022,
ytick style={color=black}
]
\addplot [black]
table {%
0.775 21139.807584635
1.225 21139.807584635
1.225 21612.5837924021
0.775 21612.5837924021
0.775 21139.807584635
};
\addplot [black]
table {%
1 21139.807584635
1 20860.9814907967
};
\addplot [black]
table {%
1 21612.5837924021
1 21920.0634416506
};
\addplot [black]
table {%
0.8875 20860.9814907967
1.1125 20860.9814907967
};
\addplot [black]
table {%
0.8875 21920.0634416506
1.1125 21920.0634416506
};
\addplot [black, mark=*, mark size=1, mark options={solid,fill opacity=0}, only marks]
table {%
1 20738.5417460026
1 20839.9397654044
1 20856.5059515475
1 21996.5614759269
1 21968.5070052745
1 21990.2225332568
};
\addplot [black]
table {%
1.775 22059.4470186977
2.225 22059.4470186977
2.225 22473.122682593
1.775 22473.122682593
1.775 22059.4470186977
};
\addplot [black]
table {%
2 22059.4470186977
2 21757.4028909473
};
\addplot [black]
table {%
2 22473.122682593
2 22626.8910466554
};
\addplot [black]
table {%
1.8875 21757.4028909473
2.1125 21757.4028909473
};
\addplot [black]
table {%
1.8875 22626.8910466554
2.1125 22626.8910466554
};
\addplot [black, mark=*, mark size=1, mark options={solid,fill opacity=0}, only marks]
table {%
2 21443.471584947
2 21615.615941991
2 21677.6005555628
2 22745.084339694
2 22771.608542854
2 22743.7358730908
};
\addplot [black]
table {%
2.775 21125.5679603179
3.225 21125.5679603179
3.225 21602.3145703083
2.775 21602.3145703083
2.775 21125.5679603179
};
\addplot [black]
table {%
3 21125.5679603179
3 20836.746290853
};
\addplot [black]
table {%
3 21602.3145703083
3 21905.5372767133
};
\addplot [black]
table {%
2.8875 20836.746290853
3.1125 20836.746290853
};
\addplot [black]
table {%
2.8875 21905.5372767133
3.1125 21905.5372767133
};
\addplot [black, mark=*, mark size=1, mark options={solid,fill opacity=0}, only marks]
table {%
3 20721.7141578917
3 20818.2250588789
3 20835.6659357181
3 21979.7633283449
3 21949.2587973926
3 21977.0702310553
};
\addplot [black]
table {%
3.775 21905.6622268007
4.225 21905.6622268007
4.225 22337.0005834836
3.775 22337.0005834836
3.775 21905.6622268007
};
\addplot [black]
table {%
4 21905.6622268007
4 21572.2622905714
};
\addplot [black]
table {%
4 22337.0005834836
4 22559.1811097979
};
\addplot [black]
table {%
3.8875 21572.2622905714
4.1125 21572.2622905714
};
\addplot [black]
table {%
3.8875 22559.1811097979
4.1125 22559.1811097979
};
\addplot [black, mark=*, mark size=1, mark options={solid,fill opacity=0}, only marks]
table {%
4 21345.2891621458
4 21499.3899095591
4 21557.6050997407
4 22631.3565580046
4 22660.530864196
4 22583.0947053783
};
\addplot [color0]
table {%
0.775 21472.8128987039
1.225 21472.8128987039
};
\addplot [color1, mark=triangle*, mark size=3, mark options={solid}, only marks]
table {%
1 21399.8591543736
};
\addplot [color0]
table {%
1.775 22270.5159320489
2.225 22270.5159320489
};
\addplot [color1, mark=triangle*, mark size=3, mark options={solid}, only marks]
table {%
2 22238.2153212795
};
\addplot [color0]
table {%
2.775 21448.1782671732
3.225 21448.1782671732
};
\addplot [color1, mark=triangle*, mark size=3, mark options={solid}, only marks]
table {%
3 21381.8495788267
};
\addplot [color0]
table {%
3.775 22122.1013573455
4.225 22122.1013573455
};
\addplot [color1, mark=triangle*, mark size=3, mark options={solid}, only marks]
table {%
4 22107.3231855715
};
\end{axis}

\end{tikzpicture}
\caption{\textbf{Real-World Experiment (CRITEO-UPLIFT1)}. $\Delta$AUUC (higher is better) on test for two \Teffect~estimators and corresponding \OurEstimator~estimators. Box plots computed on 51 random splits, whiskers at 5/95 percentiles. Note the higher $\Delta$AUUC and reduced variance of \OurEstimator~estimators.}
\label{fig:AUUC_real_dataset}
\end{center}
\vspace{.15in}
\end{figure}
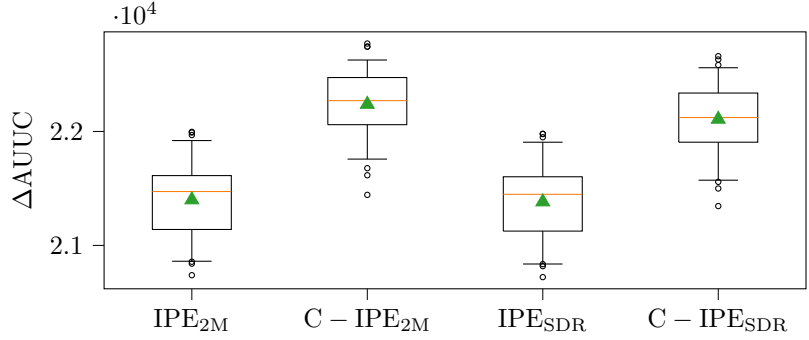

\paragraph{Real-world Experiment (CRITEO-UPLIFT1)}
To qualify the benefit of \OurEstimator~versus standard \Teffect~for real-world applications we report $\Delta$AUUC on the CRITEO-UPLIFT1 dataset. We study four methods: two \Teffect~estimators (\tmreg, and \sdrreg) and their compliance-aware variants (\tmmed, and \sdrmed). For the additional $\hat{\gamma}(x)$ model, care is taken to weight the LLH as there is a high imbalance between $\Display=1$ and $\Display=0$ classes. We use as an hyper-parameter grid the Cartesian product of \{L1, L2\} (regularization) and $\{0.01, 1, 10^2, 10^5\}$ ($C$, inverse of regularization strength) for $\hat{\compl}(\user)$. Best hyper-parameters found are L1 regularization and $C=100$.
Results are presented on Figure~\ref{fig:AUUC_real_dataset}.

The \OurEstimator~version of each estimator has a lower $\Delta$AUUC variance. This was expected from Proposition~\ref{prop:bound} and Remark~\ref{rem:bound}. Moreover, although the confidence intervals are slightly superposed, \OurEstimator~estimators outperform their \Teffect~counterparts on \textbf{each of the 51 splits}.

\paragraph{Real-world Experiment (Private Datasets)}
For the private datasets, we repeat the same procedure and compare the same metrics. Results (given in appendix) again highlight the fact that the \OurEstimator~estimators perform significantly better than their \Teffect~counterparts.

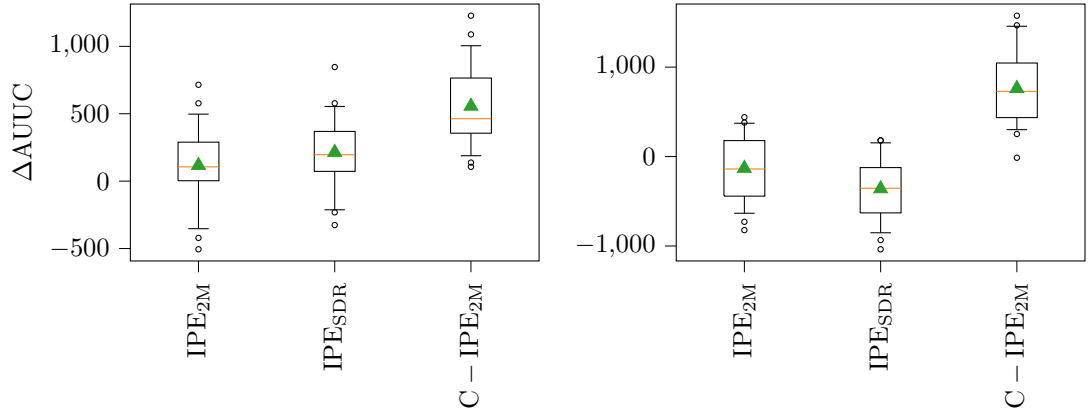
\begin{figure}[!ht]
\vspace{.15in}
\begin{center}
\pgfplotsset{width=0.45\columnwidth,height=5cm}
\begin{tikzpicture}

\definecolor{color0}{rgb}{1,0.498039215686275,0.0549019607843137}
\definecolor{color1}{rgb}{0.172549019607843,0.627450980392157,0.172549019607843}

\begin{axis}[
tick align=outside,
tick pos=left,
x grid style={white!69.0196078431373!black},
xmin=0.5, xmax=3.5,
xtick style={color=black},
xtick={1,2,3},
xticklabels={\tmreg,\sdrreg,\tmmed},
x tick label style={rotate=90,anchor=east},
y grid style={white!69.0196078431373!black},
ylabel={$\Delta$AUUC},
ymin=-592.163026769924, ymax=1313.91300943685,
ytick style={color=black}
]
\addplot [black]
table {%
0.85 3.01201296382851
1.15 3.01201296382851
1.15 289.095538040429
0.85 289.095538040429
0.85 3.01201296382851
};
\addplot [black]
table {%
1 3.01201296382851
1 -352.864961068605
};
\addplot [black]
table {%
1 289.095538040429
1 497.029491803974
};
\addplot [black]
table {%
0.925 -352.864961068605
1.075 -352.864961068605
};
\addplot [black]
table {%
0.925 497.029491803974
1.075 497.029491803974
};
\addplot [black, mark=*, mark size=1, mark options={solid,fill opacity=0}, only marks]
table {%
1 -421.141256158693
1 -505.523206942343
1 714.820457706607
1 577.500441976429
};
\addplot [black]
table {%
1.85 72.48030672869
2.15 72.48030672869
2.15 368.785983732669
1.85 368.785983732669
1.85 72.48030672869
};
\addplot [black]
table {%
2 72.48030672869
2 -212.509432272447
};
\addplot [black]
table {%
2 368.785983732669
2 554.056240640116
};
\addplot [black]
table {%
1.925 -212.509432272447
2.075 -212.509432272447
};
\addplot [black]
table {%
1.925 554.056240640116
2.075 554.056240640116
};
\addplot [black, mark=*, mark size=1, mark options={solid,fill opacity=0}, only marks]
table {%
2 -231.518393028048
2 -325.963085936955
2 846.306948886316
2 577.6373643417
};
\addplot [black]
table {%
2.85 355.525429159787
3.15 355.525429159787
3.15 764.778024959038
2.85 764.778024959038
2.85 355.525429159787
};
\addplot [black]
table {%
3 355.525429159787
3 188.596713644376
};
\addplot [black]
table {%
3 764.778024959038
3 1004.67879700594
};
\addplot [black]
table {%
2.925 188.596713644376
3.075 188.596713644376
};
\addplot [black]
table {%
2.925 1004.67879700594
3.075 1004.67879700594
};
\addplot [black, mark=*, mark size=1, mark options={solid,fill opacity=0}, only marks]
table {%
3 137.70498203312
3 106.037886311148
3 1227.27318960927
3 1088.46973885172
};
\addplot [color0]
table {%
0.85 106.014081436485
1.15 106.014081436485
};
\addplot [color1, mark=triangle*, mark size=3, mark options={solid}, only marks]
table {%
1 116.762068376201
};
\addplot [color0]
table {%
1.85 195.958554241161
2.15 195.958554241161
};
\addplot [color1, mark=triangle*, mark size=3, mark options={solid}, only marks]
table {%
2 213.324303135403
};
\addplot [color0]
table {%
2.85 462.990859380147
3.15 462.990859380147
};
\addplot [color1, mark=triangle*, mark size=3, mark options={solid}, only marks]
table {%
3 556.010578018789
};
\end{axis}

\end{tikzpicture}
\pgfplotsset{width=0.45\columnwidth,height=5cm}
\begin{tikzpicture}

\definecolor{color0}{rgb}{1,0.498039215686275,0.0549019607843137}
\definecolor{color1}{rgb}{0.172549019607843,0.627450980392157,0.172549019607843}

\begin{axis}[
tick align=outside,
tick pos=left,
x grid style={white!69.0196078431373!black},
xmin=0.5, xmax=3.5,
xtick style={color=black},
xtick={1,2,3},
xticklabels={\tmreg,\sdrreg,\tmmed},
x tick label style={rotate=90,anchor=east},
y grid style={white!69.0196078431373!black},
ymin=-1167.51212750749, ymax=1705.6339410202,
ytick style={color=black}
]
\addplot [black]
table {%
0.85 -442.457345379233
1.15 -442.457345379233
1.15 178.448316634421
0.85 178.448316634421
0.85 -442.457345379233
};
\addplot [black]
table {%
1 -442.457345379233
1 -633.953346338212
};
\addplot [black]
table {%
1 178.448316634421
1 372.516762175846
};
\addplot [black]
table {%
0.925 -633.953346338212
1.075 -633.953346338212
};
\addplot [black]
table {%
0.925 372.516762175846
1.075 372.516762175846
};
\addplot [black, mark=*, mark size=1, mark options={solid,fill opacity=0}, only marks]
table {%
1 -822.730375223926
1 -729.535738202584
1 380.192584821555
1 440.577273798037
};
\addplot [black]
table {%
1.85 -629.920393154886
2.15 -629.920393154886
2.15 -123.451653219343
1.85 -123.451653219343
1.85 -629.920393154886
};
\addplot [black]
table {%
2 -629.920393154886
2 -852.219907060615
};
\addplot [black]
table {%
2 -123.451653219343
2 153.399196914475
};
\addplot [black]
table {%
1.925 -852.219907060615
2.075 -852.219907060615
};
\addplot [black]
table {%
1.925 153.399196914475
2.075 153.399196914475
};
\addplot [black, mark=*, mark size=1, mark options={solid,fill opacity=0}, only marks]
table {%
2 -933.988458991256
2 -1036.91457893805
2 181.312959731775
2 180.902980281216
};
\addplot [black]
table {%
2.85 435.598160845878
3.15 435.598160845878
3.15 1045.82481351679
2.85 1045.82481351679
2.85 435.598160845878
};
\addplot [black]
table {%
3 435.598160845878
3 300.829388200538
};
\addplot [black]
table {%
3 1045.82481351679
3 1457.05962671548
};
\addplot [black]
table {%
2.925 300.829388200538
3.075 300.829388200538
};
\addplot [black]
table {%
2.925 1457.05962671548
3.075 1457.05962671548
};
\addplot [black, mark=*, mark size=1, mark options={solid,fill opacity=0}, only marks]
table {%
3 252.290327874166
3 -13.5456184613899
3 1575.03639245076
3 1467.54431021151
};
\addplot [color0]
table {%
0.85 -138.971113887051
1.15 -138.971113887051
};
\addplot [color1, mark=triangle*, mark size=3, mark options={solid}, only marks]
table {%
1 -131.436913299913
};
\addplot [color0]
table {%
1.85 -354.418726508061
2.15 -354.418726508061
};
\addplot [color1, mark=triangle*, mark size=3, mark options={solid}, only marks]
table {%
2 -360.239052648615
};
\addplot [color0]
table {%
2.85 729.125481722403
3.15 729.125481722403
};
\addplot [color1, mark=triangle*, mark size=3, mark options={solid}, only marks]
table {%
3 760.891374102849
};
\end{axis}

\end{tikzpicture}
\end{center}
\vspace{.15in}
\caption{\textbf{Real-world Experiment (Private Datasets)}. AUUC (higher is better) of two \Teffect~models and one \OurEstimator~model (based on two models) from which we subtracted the AUUC of a random model. Box plots are done on $33$ bootstraps, whiskers at 5/95 percentiles.}
\label{Private_AUUC}
\end{figure}

Figure \ref{Private_AUUC} illustrates the differences in $\Delta$AUUC of the learned models and a random model on $33$ bootstraps. The \OurEstimator~models performs better than the two \Teffect~models by having an AUUC significantly better than the random model, when the two \Teffect~models do not perform better than the random model. An interesting finding of this experiment is that, in practice, the mild expected benefit of \OurEstimator, predicted by the theory (because of a higher compliance than in CRITEO-UPLIFT1 dataset) does not seem discernible. This is however only an indication as there are multiple differences between the two datasets that might explain such a behavior. 

\section{Conclusion and Future Work}

We propose a novel approach on individual prescription effect (\Teffect~) estimation exploiting observed compliance to the treatment prescription.\\
Using the structural causal model framework, we define assumptions under which the \Teffect~can be expressed as a product of the individual treatment effect (\Meffect) and the individual level of compliance. 
In this setting, our post-mediation individual treatment effect (\OurEstimator) estimator is consistent. Moreover its asymptotic variance improves with the level of compliance.
Experimentally, we show how the performance of several baseline \Teffect~estimators improve when plugged in the \OurEstimator~meta-estimator. We also observe the relationship between performance and compliance as predicted by our theoretical results.\\
Finally, this work opens several perspectives among which: i) stability of our results under variations of assumptions, ii) bound tightness and properties in high-dimensional contexts, and iii) exploration of how representation learning approaches may uncover by themselves \OurEstimator-like estimator decomposition under weaker causal assumptions.

\newpage

\bibliographystyle{icml2020}
\bibliography{bib_postexpo}

\newpage
\newpage

\appendix
\section{Proofs}
To ensure a comfortable appendix read we quickly remind, for each lemma and proposition, vocabulary and notations introduced in the main text that are used in the associated proofs.
\subsection{Proposition~\ref{prop:assumptions}}
The structural causal model $\SCM = (\mathbb{S},\mathbb{P}_\mathbf{N})$ is assumed to represent the causal mechanisms underlying the variables of interest in our work.\\
$\mathbb{S}$ is defined in Equations~\eqref{eq:SCMdef}, and $\mathbb{P}_\mathbf{N}$ satisfies the following mild conditions: $N_\Conf, N_\User, N_\Display, N_\Outcome$ are noise consistent with variables definitions, and $\tilde{N}_\Treat$ is distributed according to a Bernoulli distribution with parameter $p = \Probc{\SCM}{\Treat}$, consistent with a randomized controlled experiment setting.

In Proposition~\ref{prop:assumptions}, we list four assumptions implied by the $\SCM$ about the variables of interest.

\noindent\textbf{Proof}.\\
The proof of Proposition \ref{prop:assumptions} relies on the notions of valid adjustment set (Definition~6.38 of \cite{peters2017elements}), its relation to the back-door criterion (Proposition~6.41 of \cite{peters2017elements}), the notion of d-separation (Definition 6.1 of \cite{peters2017elements}) and the Markov property (Proposition 6.21 of \cite{peters2017elements}).\\
First of all, the SCM $\SCM$ is Markovian with respect to its own entailed distribution (Proposition 6.31 of \cite{peters2017elements}): this implies that every conditional independence encoded in the graph $\mathcal{G}_\SCM$ holds in distribution $\mathbb{P}^\SCM$.

\noindent\textbf{\textit{Randomized prescription}}, is implied by the Markov property and the fact that $\Treat$ and $\User$ are d-separated by the empty set in $\mathcal{G}_\SCM$: all paths between $\Treat$ and $\User$ contain either $ \rightarrow \Display \leftarrow$ or $ \rightarrow \Outcome \leftarrow$, which are blocked by not including neither $\Display$ nor $\Outcome$.\\
\noindent\textbf{\textit{Exclusive treatment effect}} is implied by the Markov property and the fact that the set $\{\User,\Display\}$ d-separates $\Treat$ and $\Outcome$ in $\mathcal{G}_\SCM$. This is shown by listing all paths between $\Treat$ and $\Outcome$ and observing that they are all blocked by the set $\{\User,\Display\}$.\\
\noindent\textbf{\textit{One-sided non-compliance}} is straightforwardly implied by the structural assignment of $\Display$ given in Equations~\eqref{eq:SCMdef}.\\
\noindent\textbf{\textit{Valid covariate adjustment}} relies on the back-door criterion for valid adjustment sets. We remark that $\{\User\}$ satisfies the back-door criterion for ($\Display$,$\Outcome$) because (i) it is not a descendant of $\Display$ and (ii) it blocks all paths from $\Display$ to $\Outcome$ that enter $\Outcome$ through the backdoor: it is therefore a valid adjustment set for the ordered pair $(\Display, \Outcome)$.\\
\null \hfill $\blacksquare$

\subsection{Lemma~\ref{lemma1}}
The proposed method exploits the mediation variable $\Display$, \ie the treatment intake, by splitting the prescription to outcome path into a product of two \emph{subpaths}, both with a higher noise-to-signal ratio. In particular, based on the causal graphical model displayed in Figure~\ref{figure:SCM}, we can insert $\Display$ into the $\Probdocgiv{\SCM}{\Treat}{\Outcome}{\user}$ as presented in Lemma~\ref{lemma1}.

\newpage 
\noindent\textbf{Proof}.\\
Assuming the SCM $\SCM$ truly describes the relationships between $\Treat, \User,  \Display, \Outcome$, we have:
\begin{align*}
    \Probdocgiv{\SCM}{\Treat}{\Outcome}{\user} &= \Probdocgiv{\SCM}{\Treat}{\Outcome, \Display}{\user} + \Probdocgiv{\SCM}{\Treat}{\Outcome, \no{\Display}}{\user}\\
    &= \underbrace{\Probdocgiv{\SCM}{\Treat}{\Outcome}{\user, \Display}}_{\Probcgiv{\SCM}{\Outcome}{\user, \Display}}\Probdocgiv{\SCM}{\Treat}{\Display}{\user} + \underbrace{\Probdocgiv{\SCM}{\Treat}{\Outcome}{\user, \no{\Display}}}_{\Probcgiv{\SCM}{\Outcome}{\user, \no{\Display}}}\Probdocgiv{\SCM}{\Treat}{\no{\Display}}{\user} \\
    &= \Probcgiv{\SCM}{\Outcome}{\user, \Display}\Probdocgiv{\SCM}{\Treat}{\Display}{\user} + \Probcgiv{\SCM}{\Outcome}{\user, \no{\Display}}\Probdocgiv{\SCM}{\Treat}{\no{\Display}}{\user}\\
    &= \Probcgiv{\SCM}{\Outcome}{\user, \Display}\Probdocgiv{\SCM}{\Treat}{\Display}{\user} + \Probcgiv{\SCM}{\Outcome}{\user, \no{\Display}}\underbrace{\Probdocgiv{\SCM}{\Treat}{\no{\Display}}{\user}}_{1 - \Probdocgiv{\SCM}{\Treat}{\Display}{\user}} \\
    &= \Probcgiv{\SCM}{\Display}{\user, \Treat}\Big(\Probcgiv{\SCM}{\Outcome}{\user, \Display} - \Probcgiv{\SCM}{\Outcome}{\user, \no{\Display}}\Big) + \Probcgiv{\SCM}{\Outcome}{\user, \no{\Display}},
\end{align*}
where we used assumptions described Equation~\ref{eq:assumptions}, namely:
\begin{itemize}
    \item $\Probdocgiv{\SCM}{\Display}{\Outcome}{\user} = \Probcgiv{\SCM}{\Outcome}{\Display, \user}$ (\emph{Valid covariate adjustment}),
    \item $\Probdocgiv{\SCM}{\Treat}{\cdot}{\user, \cdot} = \Probcgiv{\SCM}{\cdot}{\user, \Treat, \cdot}$ (\emph{Randomized prescription}),
    \item $\Probcgiv{\SCM}{\Outcome}{\user, \Treat, \Display} = \Probcgiv{\SCM}{\Outcome}{\user, \Display}$ (\emph{Exclusive treatment effect}),
\end{itemize}
and the claim follows.
\null \hfill $\blacksquare$

\subsection{Proposition~\ref{prop:decompITE}}
For all $\user \in \Userset$, we define the individual treatment prescription effect $\TE{\Teffect}(\user)$, treatment intake effect $\TE{\Meffect}(\user)$, as well as the individual compliance $\compl(\user)$ probability in Equations~\eqref{eq:notations}.

In Proposition~\ref{prop:decompITE}, we propose a result linking the \Teffect, the \Meffect~and the compliance probability, namely:
\begin{equation*}
\TE{\Teffect}(\user) = \TE{\Meffect}(\user)\compl(\user).
\end{equation*}

\noindent\textbf{Proof}\\
We have an analogous version of \eqref{decompConversionRate} for the term $\Probdocgiv{\SCM}{\no{\Treat}}{\Outcome}{\user}$:
\begin{equation*}
\Probdocgiv{\SCM}{\no{\Treat}}{\Outcome}{\user} = \Probcgiv{\SCM}{\Outcome}{\user, \no{\Display}} + \Probcgiv{\SCM}{\Display}{\user, \no{\Treat}}\Big(\Probcgiv{\SCM}{\Outcome}{\user, \Display} - \Probcgiv{\SCM}{\Outcome}{\user, \no{\Display}}\Big).
\end{equation*}
Since $\no{\Treat} \Rightarrow \no{\Display}$ (\emph{One-sided non-compliance} assumption), we get that, $\forall \user \in \Userset$, $\Probcgiv{\SCM}{\Display}{\user, \no{\Treat}} = 0$, and finally:
$$
\Probdocgiv{\SCM}{\no{\Treat}}{\Outcome}{\user} = \Probcgiv{\SCM}{\Outcome}{\user, \no{\Display}}.
$$
Then:
\begin{align*}
    \TE{\Teffect}(\user) &= \Probdocgiv{\SCM}{\Treat}{\Outcome}{\user} - \Probdocgiv{\SCM}{\no{\Treat}}{\Outcome}{\user}\\
    &= \Probcgiv{\SCM}{\Display}{\user, \Treat}\Big(\Probcgiv{\SCM}{\Outcome}{\user, \Display} - \Probcgiv{\SCM}{\Outcome}{\user, \no{\Display}}\Big) + \Probcgiv{\SCM}{\Outcome}{\user, \no{\Display}} - \underbrace{\Probcgiv{\SCM}{\Outcome}{\user, \no{\Treat}}}_{\Probcgiv{\SCM}{\Outcome}{\user, \no{\Display}}}\\
    &= \Probcgiv{\SCM}{\Display}{\user, \Treat}\Big(\Probcgiv{\SCM}{\Outcome}{\user, \Display} - \Probcgiv{\SCM}{\Outcome}{\user, \no{\Display}}\Big).
\end{align*}
which completes the proof.\\
\null\hfill $\blacksquare$

\subsection{Proposition~\ref{prop:bound}}
\noindent\textbf{Single-stratum setting}. We focus on the \Teffect~estimation for a single value $\user_0$ of $\User$, for which we assume to observe $n$ \textit{i.i.d.} samples $\{(\user_0, \Treat_i, \Display_i, \Outcome_i)\}_{1\leq i \leq n}$. 

\noindent\textbf{Notations}. Consistently with notations presented in Equations~\ref{eq:notations}, $\alpha(\user_0)$, $\beta(\user_0)$ refer respectively to the relative \Teffect~and relative \Meffect~in stratum $\{\User = \user_0\}$ (and are assumed to be positive in this illustrative setting), and we denote $\complE(\user_0)$, $\TEE{\Teffect}(\user_0)$ and $\TEE{\Meffect}(\user_0)$ respectively the maximum-likelihood estimator (MLE) of $\compl(\user_0)$, and the MLE-based two-model estimators (difference of two MLE estimators) of $\TE{\Teffect}(\user_0)$ and $\TE{\Meffect}(\user_0)$. We define the associated \OurEstimator~estimator as $\TEE{\OurEstimator}(\user_0) \triangleq \complE(\user_0)\TEE{\Meffect}(\user_0)$. Lastly, we denote $p_1(\user_0) = \Probcgiv{\SCM}{\Outcome}{\Treat, \user_0}$.\\
In Proposition~\ref{prop:bound}, we present an asymptotic bound for the ratio of the standard deviation ($\mathrm{sd}$) of \OurEstimator~and \Teffect~estimators (additionally assuming $\Probc{\SCM}{\Treat} = \frac{1}{2}$ for simplicity), namely:
\begin{equation*}
\lim_{n\to\infty} \frac{\mathrm{sd}(\TEE{\OurEstimator}(\user_0))}{\mathrm{sd}(\TEE{\Teffect}(\user_0))} \leq \sqrt{\left(\frac{2(1 + \beta(\user_0))}{(1 - p_1(\user_0))(2 + \alpha(\user_0))}\right)\compl(\user_0)}
\end{equation*}

\noindent\textbf{Proof}.\\
The proof is splitted in four steps:
\begin{enumerate}
    \item Maximum-Likelihood and treatment effect estimators
    \item Variance of estimators derivation
    \item Variance upper and lower bounds
    \item Wrap up
\end{enumerate}
Every random quantity is henceforth implicitly considered to be `with respect to $\user_0$'.\\

\noindent\textbf{\textsc{1. Maximum-Likelihood and treatment effect estimators}}
We remind that for two categorical random variables $X$ and $Y$, for which we observe $n$ \textit{i.i.d.} samples $\{(X_i, Y_i)\}_{1\leq i \leq n}$, we have the following Maximum-Likelihood estimators for $p(x)=P(X=x)$, $p(x,y)=P(X=x, Y=y)$ and $p(y|x)=P(X=x|Y=y)$:
\begin{align}
\label{eqs:MLEgeneral}
\begin{split}
    \hat{p}(x) &= \frac{1}{n}\sumionen \mathbb{I}_{X_i = x} \\
    \hat{p}(x, y) &= \frac{1}{n}\sumionen \mathbb{I}_{X_i = x}\mathbb{I}_{Y_i = y}\\
    \hat{p}(y|x) &= \frac{\sumionen \mathbb{I}_{X_i = x}\mathbb{I}_{Y_i = y}}{\sumionen \mathbb{I}_{X_i = x}}.
    \end{split}
\end{align}
Since $\sumionen \mathbb{I}_{X_i = x} = 0 \Rightarrow \sumionen \mathbb{I}_{X_i = x}\mathbb{I}_{Y_i = y} = 0$, we adopt the convention $\hat{P}(X=x|Y=y) = 0$ in case of null denominator for a ratio estimator such as $\hat{p}(X=x|Y=y)$, ensuring they are well defined for any samples.\\

In our context, we have $n$ \textit{i.i.d.} samples $\{(\Treat_i, \Display_i, \Outcome_i)\}_{1\leq i \leq n}$ of variables $(\Treat, \Display, \Outcome)$, and that we suppose $\Probc{\SCM}{\Treat} = \frac{1}{2}$.\\
We first define the following compact notations:
\begin{align*}
    t &= \Probc{\SCM}{\Treat} &\compl = \Probcgiv{\SCM}{\Display}{\Treat}\\
    p_0 &= \Probcgiv{\SCM}{\Outcome}{\no{\Treat}} & p_1 = \Probcgiv{\SCM}{\Outcome}{\Treat}\\
    q_0 &= \Probcgiv{\SCM}{\Outcome}{\no{\Display}} &q_1 = \Probcgiv{\SCM}{\Outcome}{\Display}\\
\end{align*}

Associated MLEs $\hat{p_0}$, $\hat{p_1}$, $\hat{q_0}$, $\hat{q_1}$, $\hat{t}$ and $\hat{\compl}$ are given by Equations \eqref{eqs:MLEgeneral}. For instance:
\begin{align*}
\label{eqs:MLEs}
        \hat{t} &= \frac{1}{n}\sumionen \Treat_i &\hat{\compl} = \frac{1}{\sumionen\Treat_i}\sum\limits_{i = 1}^n \Display_i \\ 
        \hat{p}_0 &= \frac{1}{\sum\limits_{i = 1}^n (1 - \Treat_i)}\sumionen (1- \Treat_i) \Outcome_i &\hat{p}_1 = \frac{1}{\sum\limits_{i = 1}^n \Treat_i}\sumionen \Treat_i \Outcome_i \\ 
        \hat{q}_0 &= \frac{1}{\sumionen (1 - \Display_i)}\sumionen (1- \Display_i) \Outcome_i &\hat{q}_1 = \frac{1}{\sumionen \Display_i}\sumionen \Display_i \Outcome_i    
\end{align*}

Direct estimators for $\TE{\Teffect}$, $\TE{\Meffect}$ are given by applying the two-model approach to MLEs given in these equations, \ie
\begin{align*}
    \TEE{\Teffect} &= \hat{p}_1 - \hat{p}_0 \\
    \TEE{\Meffect} &= \hat{q}_1 - \hat{q}_0
\end{align*}
and the corresponding $\TE{\OurEstimator}$ estimator therefore writes:
$$
\TEE{\OurEstimator} = (\hat{q}_1 - \hat{q}_0)\hat{\compl}.
$$
In what follows, we will now write $\sumi$ instead of $\sumionen$ when there is no ambiguity.\\

\noindent\textbf{\textsc{2. Variance of estimators derivation}}\\

\noindent\textbf{\textsc{2.a. $\TEE{\Teffect}$ variance derivation}}

For any random variables $X, Y$, we remind that:
\begin{equation}\label{generalVarianceDecomp}
\mathrm{Var}(X) = \mathrm{Var}(\E{X|Y}) + \E{\mathrm{Var}(X|Y)}.
\end{equation}
Which applied with $X = \TEE{\Teffect} = \hat{p}_1 - \hat{p}_0$ and $Y = \{\Treat_k\}_k := \{\Treat_1, \dots, \Treat_n\}$, gives:
\begin{equation}
\label{decompositionVarianceITE}
\mathrm{Var}(\TEE{\Teffect}) = \underbrace{\mathbb{E}\left[\mathrm{Var}(\TEE{\Teffect} | \{\Treat_k\}_k)\right]}_{\circled{A}} + \underbrace{\mathrm{Var}\left[\mathbb{E}(\TEE{\Teffect}|\{\Treat_k\}_k)\right]}_{\circled{B}}.    
\end{equation}

\noindent\textbf{Computation of \circled{A}}\\
The term $\mathrm{Var}(\TEE{\Teffect} | \{\Treat_k\}_k)$ decomposes as:
\begin{equation}
\label{eq:intermediary_A}
    \mathrm{Var}(\TEE{\Teffect} | \{\Treat_k\}_k) = \underbrace{\mathrm{Var}(\hat{p}_1 | \{\Treat_k\}_k)}_{\circled{A1}} + \underbrace{\mathrm{Var}(\hat{p}_0 | \{\Treat_k\}_k)}_{\circled{A2}} - 2\underbrace{\mathrm{Cov}(\hat{p}_1, \hat{p}_0 | \{\Treat_k\}_k)}_{\circled{A3}}.
\end{equation}

For \circled{A1}, we write:
\begin{align*}
    \mathrm{Var}(\hat{p}_1 | \{\Treat_k\}_k) &= \mathrm{Var}\left(\frac{1}{\sumi \Treat_i} \sumi \Treat_i\Outcome_i \given \{\Treat_k\}_k\right) \\
    &= \left(\frac{1}{\sumi \Treat_i}\right)^2 \mathrm{Var}\left(\sumi \Treat_i\Outcome_i \given \{\Treat_k\}_k\right)\\
    &= \left(\frac{1}{\sumi \Treat_i}\right)^2 \sumi \mathrm{Var}(\Treat_i\Outcome_i \given \underbrace{\{\Treat_k\}_k}_{\Treat_i \text{ (since \textit{i.i.d.})}})\\
    &= \left(\frac{1}{\sumi \Treat_i}\right)^2 \sumi \underbrace{\Treat_i \mathrm{Var}(\Outcome_i | \Treat_i)}_{\mathrm{Var}(\Outcome \given \Treat = 1)\Treat_i\text{ (since \textit{i.i.d.})}}\\
    &= \left(\frac{1}{\sumi \Treat_i}\right)^2 \mathrm{Var}(\Outcome\given\Treat = 1) \sumi \Treat_i\\
    &= \frac{1}{\sumi \Treat_i} \mathrm{Var}(\Outcome\given\Treat = 1)\\
    &= \frac{1}{\sumi \Treat_i} p_1(1-p_1).
\end{align*}
For \circled{A2}, we analogously get:
$$
\mathrm{Var}(\hat{p}_0 | \{\Treat_k\}_k) = \frac{1}{\sumi (1 - \Treat_i)} p_0(1-p_0).
$$
For \circled{A3}, using the bi-linearity of covariance, we write:
\begin{align*}
    \mathrm{Cov}\left(\hat{p}_1, \hat{p}_0 | \{\Treat_k\}_k\right) &= \mathrm{Cov}\left(\frac{1}{\sumi \Treat_i}\sumi \Treat_i\Outcome_i, \frac{1}{\sumj (1- \Treat_j)} \sumj (1-\Treat_j)\Outcome_j | \{\Treat_k\}_k\right) \\
    &= \frac{1}{\sumi \Treat_i}\frac{1}{\sumj (1- \Treat_j)}\mathrm{Cov}\left(\sumi \Treat_i\Outcome_i, \sumj (1-\Treat_j)\Outcome_j | \{\Treat_k\}_k\right)\\
    &= \frac{1}{\sumi \Treat_i}\frac{1}{\sumj (1- \Treat_j)}\sumi \sumj (1-\Treat_j) \Treat_i \underbrace{\mathrm{Cov}\left( \Outcome_i,  \Outcome_j | \{\Treat_k\}_k\right)}_{\neq 0 \text{ only if }i=j \text{ (since \textit{i.i.d.})}}\\
    &= \frac{1}{\sumi \Treat_i}\frac{1}{\sumj (1- \Treat_j)}\sumi (1-\Treat_i) \Treat_i \mathrm{Cov}( \Outcome_i,  \Outcome_i | \underbrace{\{\Treat_k\}_k}_{\Treat_i \text{ (since \textit{i.i.d.})}})\\
    &= \frac{1}{\sumi \Treat_i}\frac{1}{\sumj (1- \Treat_j)}\sumi \underbrace{(1-\Treat_i) \Treat_i}_{=0} \mathrm{Var}( \Outcome_i |\Treat_i)\\
    &=0.
\end{align*}

Injecting values of \circled{A1}, \circled{A2} and \circled{A3} in Equation~\eqref{eq:intermediary_A} we get:
\begin{equation}
    \label{VarTEEITE}
    \mathrm{Var}(\TEE{\Teffect} | \{\Treat_k\}_k) = \frac{1}{\sumi \Treat_i}p_1(1-p_1) + \frac{1}{\sumi (1-\Treat_i)}p_0(1-p_0).
\end{equation}
Which leads to the following expression for \circled{A}:
\begin{equation}\label{firsttermDecompVarianceITE}
    \mathbb{E}\left[\mathrm{Var}(\TEE{\Teffect} | \{\Treat_k\}_k)\right] = \mathbb{E}\left[\frac{1}{\sumi \Treat_i}\right]p_1(1-p_1) + \mathbb{E}\left[\frac{1}{\sumi (1-\Treat_i)}\right]p_0(1-p_0).
\end{equation}

\noindent\textbf{Computation of \circled{B}}\\
We may write
\begin{align*}
    \mathbb{E}(\TEE{\Teffect}|\{\Treat_k\}_k) &= \underbrace{\mathbb{E}(\hat{p_1}|\{\Treat_k\}_k)}_{\circled{B1}} - \underbrace{\mathbb{E}(\hat{p_0}|\{\Treat_k\}_k)}_{\circled{B2}}.
\end{align*}
For \circled{B1}, we write:
\begin{align*}
\mathbb{E}(\hat{p_1}|\{\Treat_k\}_k) &= \mathbb{E}\left(\frac{1}{\sumi \Treat_i} \sumi \Treat_i \Outcome_i |\{\Treat_k\}_k\right) \\
&= \frac{1}{\sumi \Treat_i} \sumi \mathbb{E}\left(\Treat_i \Outcome_i |\{\Treat_k\}_k\right) \\
&= \frac{1}{\sumi \Treat_i} \sumi \Treat_i \underbrace{\mathbb{E}(\Outcome_i |\Treat_i)}_{p_1} \\
&= \frac{1}{\sumi \Treat_i} \left(\sumi \Treat_i\right)p_1\\
&=p_1.
\end{align*}
For \circled{B2}, we analogously get: $\mathbb{E}(\hat{p_0}|\{\Treat_k\}_k) = p_0$.\\
Therefore $\mathbb{E}(\TEE{\Teffect}|\{\Treat_k\}_k) = p_1 - p_0$ is constant relatively to $\{T_k\}_k$, and we have computed \circled{B}:
\begin{equation}
    \label{secondtermDecomVarianceITE}
    \mathrm{Var}\left[\mathbb{E}(\TEE{\Teffect}|\{\Treat_k\}_k)\right] = 0.
\end{equation}

Combining Equations~\eqref{decompositionVarianceITE}, \eqref{firsttermDecompVarianceITE} and \eqref{secondtermDecomVarianceITE} we end up with:
\begin{equation}\label{varianceITEE}
    \mathrm{Var}(\TEE{\Teffect}) = \mathbb{E}\left[\frac{1}{\sumi \Treat_i}\right]p_1(1-p_1) + \mathbb{E}\left[\frac{1}{\sumi (1-\Treat_i)}\right]p_0(1-p_0).
\end{equation}

\vspace{5mm}
\noindent\textbf{\textsc{2.b. $\TEE{\OurEstimator}$ variance derivation}}

Using Equation~\eqref{generalVarianceDecomp} with $X = \TEE{\OurEstimator} = \hat{\compl}(\hat{q}_1 - \hat{q}_0)$ and $Y = \{\Treat_k,\Display_k\}_k$ we may write:
\begin{equation}
\label{decompositionVarianceMITE}
\mathrm{Var}(\TEE{\OurEstimator}) = \underbrace{\mathbb{E}\left[\mathrm{Var}(\TEE{\OurEstimator} | \{\Treat_k,\Display_k\}_k)\right]}_{\circled{C}} + \underbrace{\mathrm{Var}\left[\mathbb{E}(\TEE{\OurEstimator}|\{\Treat_k,\Display_k\}_k)\right]}_{\circled{D}}.    
\end{equation}
\noindent\textbf{Computation of \circled{C}}\\
Using the fact that $\TEE{\OurEstimator} = \hat{\compl}(\hat{q}_1 - \hat{q}_0)$, and remarking that $\Egiv{\hat{\compl}}{\{\Treat_k,\Display_k\}_k} = \hat{\compl}$, we may write:
$$
    \mathrm{Var}(\hat{\compl}(\hat{q}_1 - \hat{q}_0) | \{\Treat_k,\Display_k\}_k) = \hat{\compl}^2 \mathrm{Var}((\hat{q}_1 - \hat{q}_0) | \{\Treat_k,\Display_k\}_k)
$$
By analogy with Equation~\eqref{VarTEEITE}, we have
$$
    \mathrm{Var}(\hat{q}_1 - \hat{q}_0) | \{\Treat_k,\Display_k\}_k) = \frac{1}{\sumi \Display_i}q_1(1-q_1) + \frac{1}{\sumi(1 - \Display_i)}q_0(1-q_0),
$$
which gives
\begin{equation}
    \label{firstTermCheckpoint}
 \mathrm{Var}(\hat{\compl}(\hat{q}_1 - \hat{q}_0) | \{\Treat_k,\Display_k\}_k) = \hat{\compl}^2(\frac{1}{\sumi \Display_i} q_1(1-q_1) + \frac{1}{\sumi(1 - \Display_i)}q_0(1-q_0)).
\end{equation}
Injecting $\hat{\compl} = \frac{\sumi \Display_i}{\sumi \Treat_i} = \frac{\sumi \Display_i\Treat_i}{\sumi \Treat_i}$
in \eqref{firstTermCheckpoint}, we get:
$$
\mathrm{Var}(\hat{\compl}(\hat{q}_1 - \hat{q}_0) | \{\Treat_k,\Display_k\}_k) = \frac{\sumi \Display_i}{\left(\sumi \Treat_i\right)^2}q_1(1-q_1) + \frac{\left(\sumi \Display_i\right)^2}{\left(\sumi \Treat_i\right)^2\sumi(1 - \Display_i)}q_0(1-q_0).
$$
Taking the expectancy, we therefore get the following expression for \circled{C}:
\begin{equation}
    \label{firstTermDecompVarianceMITE}
    \mathbb{E}\left[\mathrm{Var}(\TEE{\OurEstimator} | \{\Treat_k,\Display_k\}_k)\right] = \mathbb{E}\left[\frac{\sumi \Display_i}{\left(\sumi \Treat_i\right)^2}\right]q_1(1-q_1) + \mathbb{E}\left[\frac{\left(\sumi \Display_i\right)^2}{\left(\sumi \Treat_i\right)^2\sumi(1 - \Display_i)}\right]q_0(1-q_0).
\end{equation}

\noindent\textbf{Computation of \circled{D}}\\
Using the fact that $\hat{\compl}$ is $\{\Treat_k, \Display_k\}_k-$measurable, we get:
\begin{equation}
\label{eq:intermediary_D}
    \mathbb{E}(\TEE{\OurEstimator}|\{\Treat_k,\Display_k\}_k) 
    = \hat{\compl}(\underbrace{\mathbb{E}(\hat{q}_1|\{\Treat_k,\Display_k\}_k)}_{\circled{D1}} - \underbrace{\mathbb{E}(\hat{q}_0|\{\Treat_k,\Display_k\}_k)}_{\circled{D2}}).
\end{equation}
For \circled{D1}, a few mechanical computations bring:
\begin{align*}
    \mathbb{E}(\hat{q}_1|\{\Treat_k,\Display_k\}_k) &= \mathbb{E}\left(\frac{1}{\sumi \Display_i}\sumi \Display_i \Outcome_i|\{\Treat_k,\Display_k\}_k\right)\\
    &= \frac{1}{\sumi \Display_i}\mathbb{E}\left(\sumi \Display_i \Outcome_i|\{\Treat_k,\Display_k\}_k\right)\\
    &= \frac{1}{\sumi \Display_i} \sumi \Display_i\mathbb{E}\left( \Outcome_i|\{\Treat_k,\Display_k\}_k\right)\\
    &= \frac{1}{\sumi \Display_i} \sumi \underbrace{\Display_i\mathbb{E}\left( \Outcome_i|\Treat_i, \Display_i \right)}_{\Display_i \Egiv{\Outcome}{\Display=1}}\\
    &= \underbrace{\Egiv{\Outcome}{\Display=1}}_{q_1}\frac{1}{\sumi \Display_i} \sumi \Display_i \\
    &=q_1
\end{align*}
For \circled{D2}, we analogously get:
$$
 \mathbb{E}(\hat{q}_0|\{\Treat_k,\Display_k\}_k) = q_0.
$$
Injecting these values in \eqref{eq:intermediary_D}:
\begin{equation*}
    \mathbb{E}(\TEE{\OurEstimator}|\{\Treat_k,\Display_k\}_k) = \hat{\compl}(q_1 - q_0).
\end{equation*}
which gives the following expression for \circled{D}:
$$
    \mathrm{Var}\left(\mathbb{E}(\TEE{\OurEstimator}|\{\Treat_k,\Display_k\}_k)\right) = (q_1 - q_0)^2 \mathrm{Var}(\hat{\compl}).
$$
Using the fact that $\mathrm{Var}(\hat{\compl}) = \mathbb{E}(\mathrm{Var}(\hat{\compl} | \{\Treat_k\})) + \mathrm{Var}(\Egiv{\hat{\compl}}{\{\Treat_k\}})$, we get:
$$
\mathrm{Var}(\hat{\compl}) = \mathbb{E}\left(\frac{1}{\sumi \Treat_i}\right)\compl(1-\compl).
$$
We end up with the final expression for \circled{D}:
\begin{equation}
    \label{secondTermDecompVarianceMITE}
    \mathrm{Var}\left(\mathbb{E}\left[\TEE{\OurEstimator}|\{\Treat_k,\Display_k\}_k\right]\right) = (q_1 - q_0)^2 \mathbb{E}\left[\frac{1}{\sumi \Treat_i}\right]\compl(1-\compl).
\end{equation}

Combining Equations~\eqref{decompositionVarianceMITE}, \eqref{firstTermDecompVarianceMITE} and \eqref{secondTermDecompVarianceMITE}, we have:
\begin{equation}
    \label{varianceMITEE}
    \mathrm{Var}(\TEE{\OurEstimator}) = \mathbb{E}\left[\frac{\sumi \Display_i}{\left(\sumi \Treat_i\right)^2}\right]q_1(1-q_1) + \mathbb{E}\left[\frac{\left(\sumi \Display_i\right)^2}{\left(\sumi \Treat_i\right)^2\sumi(1 - \Display_i)}\right]q_0(1-q_0) + (q_1 - q_0)^2 \mathbb{E}\left(\frac{1}{\sumi \Treat_i}\right)\compl(1-\compl).
\end{equation}

\vspace{5mm}
\noindent\textbf{\textsc{3. Asymptotic variance upper and lower bounds}}

\noindent\textbf{\textsc{3.a Asymptotic lower bound of $\mathrm{Var}(\TEE{\Teffect})$}}\
With a slight rewriting of \eqref{varianceITEE},
$$
    n \mathrm{Var}(\TEE{\Teffect}) = \mathbb{E}\left[\frac{1}{\frac{1}{n}\sumi \Treat_i}\right]p_1(1-p_1) + \mathbb{E}\left[\frac{1}{\frac{1}{n}\sumi (1-\Treat_i)}\right]p_0(1-p_0).
$$
Then because $t = \Probc{\SCM}{\Treat}= \frac{1}{2}$, by the law of large numbers
, we get 
\begin{equation}
\label{varianceITEElimit}
        \underset{n \rightarrow \infty}{\lim} n \mathrm{Var}(\TEE{\Teffect}) = 2\left(p_1(1-p_1) + p_0(1-p_0)\right).
\end{equation}
Now, using $p_1 = (1+\alpha)p_0$, we can write:
\begin{align*}
    \underset{n \rightarrow \infty}{\lim} n \mathrm{Var}(\TEE{\Teffect}) &= 2p_0\left(1-p_0 + (1+\alpha)(1 - p_1)\right)\\
    &\geq 2p_0\left(1-p_1 + (1 + \alpha)(1-p_1)\right)\\
    &= 2p_0(1-p_1)(2+\alpha),
\end{align*}
where the first inequality uses $\alpha \geq 0$.
In summary, we have the following asymptotic lower bound for $\mathrm{Var}(\TEE{\Teffect})$:
\begin{equation}
        \label{asymptoticLowerBoundITEE}
\underset{n \rightarrow \infty}{\lim} n \mathrm{Var}(\TEE{\Teffect}) \geq 2p_0(1-p_1)(2+\alpha).
\end{equation}

\noindent\textbf{\textsc{3.a Asymptotic upper bound of $\mathrm{Var}(\TEE{\OurEstimator})$}}\\
From \eqref{varianceMITEE}, we have
\begin{multline}
    n \mathrm{Var}(\TEE{\OurEstimator}) = \mathbb{E}\left[\frac{\frac{1}{n}\sumi \Display_i}{\left(\frac{1}{n}\sumi \Treat_i\right)^2}\right]q_1(1-q_1) + \mathbb{E}\left[\frac{\left(\frac{1}{n}\sumi \Display_i\right)^2}{\left(\frac{1}{n}\sumi \Treat_i\right)^2\frac{1}{n}\sumi(1 - \Display_i)}\right]q_0(1-q_0)\\ + (q_1 - q_0)^2 \mathbb{E}\left(\frac{1}{\frac{1}{n}\sumi \Treat_i}\right)\compl(1-\compl).
\end{multline}

Again, using the law of large numbers
, we get
\begin{equation}
\label{varianceMITEElimit}
        \underset{n \rightarrow \infty}{\lim} n \mathrm{Var}(\TEE{\OurEstimator}) = 2\compl q_1 (1-q_1) + 2\frac{\compl^2}{2 - \compl} q_0 (1 - q_0) + 2\compl(1-\compl)(q_1 - q_0)^2.
\end{equation}
Now, using that for any $q \in [0,1]$, $q(1-q) \leq q$, and reminding that $q_1 = (1+\beta)q_0 \leq 1$ where $\beta \geq 0$ by assumption, we get:
\begin{align*}
    \underset{n \rightarrow \infty}{\lim} n \mathrm{Var}(\TEE{\OurEstimator}) &= 2\compl q_1 (1-q_1) + 2\frac{\compl^2}{2 - \compl} q_0 (1 - q_0) + 2\compl(1-\compl)(q_1 - q_0)\\
    &\leq 2\compl\left(\underbrace{q_1}_{\beta q_0} + \underbrace{\frac{\compl}{2-\compl}}_{\leq \compl \leq 1}q_0 + \underbrace{(q_0 (1+\beta))^2}_{\leq q_0 (1+ \beta)}\right)\\
    &\leq 2\compl q_0\left(1 + \beta + 1 + \beta \right)\\
    &= 4 q_0 \compl (1 + \beta).
\end{align*}
In summary, we have the following asymptotic upper bound for $\mathrm{Var}(\TEE{\OurEstimator})$:
\begin{equation}
    \label{asymptoticUpperBoundMITEE}
\underset{n \rightarrow \infty}{\lim} n \mathrm{Var}(\TEE{\OurEstimator}) \leq 4 q_0 \compl (1 + \beta).
\end{equation}

\noindent\textbf{\textsc{4. Wrap up}}\\
Combining Equations \eqref{asymptoticLowerBoundITEE} and \eqref{asymptoticUpperBoundMITEE} (ratio of positive values), we get
\begin{align*}
    \frac{\underset{n \rightarrow \infty}{\lim} n \mathrm{Var}(\TEE{\OurEstimator})}{\underset{n \rightarrow \infty}{\lim} n \mathrm{Var}(\TEE{\Teffect})} &\leq \frac{4 q_0 \compl (1 + \beta)}{2p_0(1-p_1)(2+\alpha)} \\
    &= 2\frac{1+\beta}{(1-p_1)(2+\alpha)}\compl. 
\end{align*}
Where we remind that \textit{One-sided non-compliance} and \textit{Exclusive treatment effect} imply straightforwardly that $p_0 = q_0$ (as shown in the beginning of the proof of Proposition \ref{prop:decompITE}).
Since the limits of both the numerator and denominator exist, this implies that
$$
\underset{n \rightarrow \infty}{\lim} \frac{\mathrm{Var}(\TEE{\OurEstimator})}{\mathrm{Var}(\TEE{\Teffect})} \leq 2\frac{1+\beta}{(1-p_1)(2+\alpha)}\compl.
$$
Taking the square root of this equation gives the wanted result.\\
\null \hfill $\blacksquare$

\section{Additional Results and Experimentation Details}

\subsection{AUUC}

AUUC \citep{rzepakowski2012decision, rzepakowski2010decision, radcliffe2011real} is the Area Under the Uplift Curve.
It is obtained by ranking the users of a test set according to their predicted uplift, in descending order. In what follows, we will focus on the evaluation of \Teffect models (effect of $\Teffect$ on $\Outcome$).

The uplift curve starts at the point $(0,0)$, then for each user (in decreasing order of the predicted uplift) it goes up of 1 point if the user is in group $\Treat = 1$ with $\Outcome=1$, it goes down of 1 point if the user is in group $\Treat = 0$ with $\Outcome=1$, and it stays flat if the user has zero outcome $(\Outcome=0)$. So the label $l$ is $\Outcome * (2\Treat -1)$. We normalise the x-axis so that it goes from $0$ to $1$.

The aim of an \Teffect model is to rank first the users with positive outcome and assigned to prescription ($\Treat = 1$), then the users with negative outcome, and finally the users with positive outcome and assigned to non-prescription ($\Treat = 0$). A model respecting such a ranking would have a maximal AUUC.

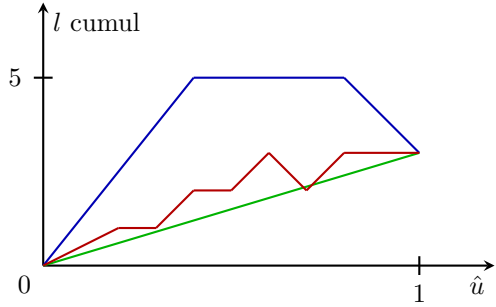
\begin{figure}[ht]
\begin{center}
\begin{tikzpicture}
[scale=0.5,
nodestyle/.style={circle,draw=black,fill=gray!10, inner sep=2pt},
edgestyle/.style={-},
>=stealth]

\draw[thick,->] (0,0) -- (12,0) node[anchor=north east] {$\hat{u}$};
\draw[thick,->](0,0) -- (0,7) node[anchor=north west] {$l$ cumul};

\draw[thick,black!30!blue] (0,0) -- (4,5) ;
\draw[thick,black!30!blue] (4,5) -- (8,5) ;
\draw[thick,black!30!blue] (8,5) -- (10,3) ;

\draw[thick,black!30!green] (0,0) -- (10,3) ;

\draw[thick,black!30!red] (0,0) -- (2,1) ;
\draw[thick,black!30!red] (2,1) -- (3,1) ;
\draw[thick,black!30!red] (3,1) -- (4,2) ;
\draw[thick,black!30!red] (4,2) -- (5,2) ;
\draw[thick,black!30!red] (5,2) -- (6,3) ;
\draw[thick,black!30!red] (6,3) -- (7,2) ;
\draw[thick,black!30!red] (7,2) -- (8,3) ;
\draw[thick,black!30!red] (8,3) -- (10,3) ;

\coordinate (A0) at (10, -.75);
\node (A0) at (A0){$1$};
\draw[thick] (10,-0.25) -- (10,0.25);

\coordinate (A1) at (-0.5,-0.5);
\node (A1) at (A1) {$0$};

\coordinate (A2) at (-.75, 5);
\node (A2) at (A2){$5$};
\draw[thick] (-0.25,5) -- (0.25,5);

\end{tikzpicture}
\caption{Example of uplift curves for (i) a perfect model (blue), a random model (green) and an example \Teffect~model (red). The AUUC of these models correspond to the area under their respective uplift curves.}
\label{fig:AUUC_ex}
\end{center}
\end{figure}

The value of the AUUC highly depends on the dataset it is computed on (the test set in the experiments presented in the main text).
The first point of the uplift curve is always $(0,0)$ and the last one is always $(1,U)$ were $U = \sum \Outcome\Treat - \sum \Outcome (1-\Treat)$.
The variability of the test set (when we do different splits such as in the experiments presented in the main text) accounts for an important part of the variability of the metric on different splits, which we handle by subtracting the (average) AUUC of a random model $-$ equal to $\frac{U}{2}$ $-$ to the AUUC of the evaluated models: this measure is called $\Delta$AUUC.

\subsection{Model training}
Our goal is to compare the performance of two approaches for the esimation of the \Teffect~(Individual Treatment Prescription Effect): (1) our proposed \OurEstimator~(compliance-aware Individual Treatment Effect) approach, which exploits observed compliance, and (2) standard \Teffect estimators which do not exploit observed compliance.
We implement two kinds of models for both \Teffect~and \Meffect~(Individual Treatment Intake Effect) factors used inside \OurEstimator estimators.
These models are (1) the Two-Models, with one model learned on the group $\Treat = 1$ and one model learned on the group $\Treat = 0$, and (2) Shared Data Representation (SDR) \citep{betlei2018uplift}, which is inspired from multi-task learning and has more capacity than the basic Two-Models approach.

\begin{figure}[ht]
\begin{center}
\pgfplotsset{width=.5\columnwidth,height=5cm}
\begin{tikzpicture}

\definecolor{color0}{rgb}{1,0.498039215686275,0.0549019607843137}
\definecolor{color1}{rgb}{0.172549019607843,0.627450980392157,0.172549019607843}

\begin{axis}[
tick align=outside,
tick pos=left,
x grid style={white!69.0196078431373!black},
xmin=0.5, xmax=6.5,
xtick style={color=black},
xtick={1,2,3,4,5,6},
xticklabels={ITE\_best,\tmreg,\tmmed,\sdrreg,\sdrmed,Oracle},
x tick label style={rotate=90,anchor=east},
y grid style={white!69.0196078431373!black},
ylabel={$\Delta$AUUC},
ymin=31.4947011408267, ymax=466.403339381875,
ytick style={color=black}
]
\addplot [black]
table {%
0.75 159.999561257999
1.25 159.999561257999
1.25 240.313842895585
0.75 240.313842895585
0.75 159.999561257999
};
\addplot [black]
table {%
1 159.999561257999
1 107.750273912996
};
\addplot [black]
table {%
1 240.313842895585
1 301.245939948381
};
\addplot [black]
table {%
0.875 107.750273912996
1.125 107.750273912996
};
\addplot [black]
table {%
0.875 301.245939948381
1.125 301.245939948381
};
\addplot [black, mark=*, mark size=1, mark options={solid,fill opacity=0}, only marks]
table {%
1 98.7428360598626
1 51.2632756063289
1 101.040659153888
1 305.206111135999
1 317.953603691599
1 345.535446778392
};
\addplot [black]
table {%
1.75 159.999561257999
2.25 159.999561257999
2.25 240.313842895585
1.75 240.313842895585
1.75 159.999561257999
};
\addplot [black]
table {%
2 159.999561257999
2 107.750273912996
};
\addplot [black]
table {%
2 240.313842895585
2 301.245939948381
};
\addplot [black]
table {%
1.875 107.750273912996
2.125 107.750273912996
};
\addplot [black]
table {%
1.875 301.245939948381
2.125 301.245939948381
};
\addplot [black, mark=*, mark size=1, mark options={solid,fill opacity=0}, only marks]
table {%
2 98.7428360598626
2 51.2632756063289
2 101.040659153888
2 305.206111135999
2 317.953603691599
2 345.535446778392
};
\addplot [black]
table {%
2.75 267.553090604389
3.25 267.553090604389
3.25 342.484030068678
2.75 342.484030068678
2.75 267.553090604389
};
\addplot [black]
table {%
3 267.553090604389
3 220.048516217658
};
\addplot [black]
table {%
3 342.484030068678
3 397.539002889244
};
\addplot [black]
table {%
2.875 220.048516217658
3.125 220.048516217658
};
\addplot [black]
table {%
2.875 397.539002889244
3.125 397.539002889244
};
\addplot [black, mark=*, mark size=1, mark options={solid,fill opacity=0}, only marks]
table {%
3 199.367853017679
3 178.31849232401
3 192.161535596492
3 407.086716059576
3 446.634764916373
3 400.674916529879
};
\addplot [black]
table {%
3.75 159.999561257999
4.25 159.999561257999
4.25 240.313842895585
3.75 240.313842895585
3.75 159.999561257999
};
\addplot [black]
table {%
4 159.999561257999
4 107.750273912996
};
\addplot [black]
table {%
4 240.313842895585
4 301.245939948381
};
\addplot [black]
table {%
3.875 107.750273912996
4.125 107.750273912996
};
\addplot [black]
table {%
3.875 301.245939948381
4.125 301.245939948381
};
\addplot [black, mark=*, mark size=1, mark options={solid,fill opacity=0}, only marks]
table {%
4 98.7428360598626
4 51.2632756063289
4 101.040659153888
4 305.206111135999
4 317.953603691599
4 345.535446778392
};
\addplot [black]
table {%
4.75 267.553090604389
5.25 267.553090604389
5.25 342.484030068678
4.75 342.484030068678
4.75 267.553090604389
};
\addplot [black]
table {%
5 267.553090604389
5 220.048516217658
};
\addplot [black]
table {%
5 342.484030068678
5 397.539002889244
};
\addplot [black]
table {%
4.875 220.048516217658
5.125 220.048516217658
};
\addplot [black]
table {%
4.875 397.539002889244
5.125 397.539002889244
};
\addplot [black, mark=*, mark size=1, mark options={solid,fill opacity=0}, only marks]
table {%
5 199.367853017679
5 178.31849232401
5 192.161535596492
5 407.086716059576
5 446.634764916373
5 400.674916529879
};
\addplot [black]
table {%
5.75 271.857892386155
6.25 271.857892386155
6.25 342.515021943608
5.75 342.515021943608
5.75 271.857892386155
};
\addplot [black]
table {%
6 271.857892386155
6 220.048516217658
};
\addplot [black]
table {%
6 342.515021943608
6 397.539002889244
};
\addplot [black]
table {%
5.875 220.048516217658
6.125 220.048516217658
};
\addplot [black]
table {%
5.875 397.539002889244
6.125 397.539002889244
};
\addplot [black, mark=*, mark size=1, mark options={solid,fill opacity=0}, only marks]
table {%
6 199.367853017679
6 186.012625008338
6 192.161535596492
6 405.604716864809
6 446.634764916373
6 400.674916529879
};
\addplot [color0]
table {%
0.75 187.749872660991
1.25 187.749872660991
};
\addplot [color1, mark=triangle*, mark size=3, mark options={solid}, only marks]
table {%
1 201.747553781137
};
\addplot [color0]
table {%
1.75 187.749872660991
2.25 187.749872660991
};
\addplot [color1, mark=triangle*, mark size=3, mark options={solid}, only marks]
table {%
2 201.747553781137
};
\addplot [color0]
table {%
2.75 310.100271039719
3.25 310.100271039719
};
\addplot [color1, mark=triangle*, mark size=3, mark options={solid}, only marks]
table {%
3 308.464807710013
};
\addplot [color0]
table {%
3.75 187.749872660991
4.25 187.749872660991
};
\addplot [color1, mark=triangle*, mark size=3, mark options={solid}, only marks]
table {%
4 201.747553781137
};
\addplot [color0]
table {%
4.75 310.100271039719
5.25 310.100271039719
};
\addplot [color1, mark=triangle*, mark size=3, mark options={solid}, only marks]
table {%
5 308.563339214084
};
\addplot [color0]
table {%
5.75 311.864905920988
6.25 311.864905920988
};
\addplot [color1, mark=triangle*, mark size=3, mark options={solid}, only marks]
table {%
6 310.291914707032
};
\end{axis}

\end{tikzpicture}
\caption{$\Delta$AUUC (higher is better) of two \Teffect~models, corresponding \OurEstimator~models and Oracle model (theoretical truth). Box plots are done on 51 random splits, whiskers at 5/95 percentiles. Note how \OurEstimator~systematically increases the AUUC of standard \Teffect~estimators}
\label{fig:AUUC}
\end{center}
\end{figure}
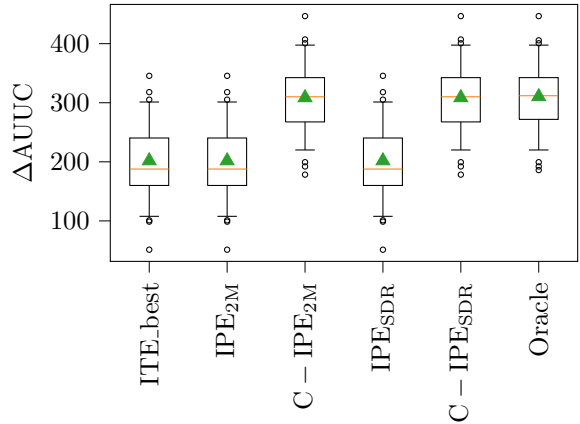

On synthetic data, we also compare the models to two theoretical models: $\mathrm{\Teffect}_{\mathrm{best}}$ and Oracle:
\begin{itemize}
    \item ${\mathrm{\Teffect}}_{\mathrm{best}}$ is the best possible model learn-able using training data but without exploiting information from variable $\Display$ (or equivalently, compliance). In short, this model predicts the difference between the empirical positive outcome rate in the group $\Treat=1$ and in the group $\Treat=0$, and does so for each user context $\user \in \Userset$. This approach is valid in the case of our synthetic dataset since we observe a high number of users for each possible context $\user$.\\
    This is indeed the best possible learnable model, because the dataset is generated with one different probability of positive outcome per category: there is no additional information any model could capture without exploiting the variable $\Display$.\\
    $\mathrm{\Teffect}_{\mathrm{best}}$ allows us to show that, in our synthetic dataset, two models and SDR perform close to the best possible \Teffect~approach. In that specific low-dimensional case, there is therefore no need to implement more complex models (Figure \ref{fig:AUUC}) such as doubly-robust methods or tree/forest-based methods.\\
    
    \item Oracle predicts the theoretical ground truth uplift (used to generate the dataset). Its AUUC represents the maximum theoretically reachable AUUC (expected). In practice, we observe that it is however not the best model on all random splits. This can be explained by the fact that the (test) dataset is randomly generated, and that the ranking of users in the test set can therefore differ slightly from the theoretical ranking.
\end{itemize}
All learned models (\Teffect, \OurEstimator~and $\mathrm{\Teffect}_{\mathrm{best}}$) suffer from the randomness of the training dataset.
 
\subsection{Model testing}

In addition of the randomness of the training dataset, the test dataset is also random.
This adds noise to the AUUC values that are computed in practice. We design a metric called, $\mathrm{AUUC}_{\mathrm{thout}}$ that computes the AUUC on a theoretical test set.
This metric may only be implemented on synthetic data.
It uses the "Theoretical Outcome" (thout) of each category of users as a label, thus circumventing the randomness inherent to the test set generation.

\begin{figure}[ht]
\begin{center}
\pgfplotsset{width=.5\columnwidth,height=5cm}
\begin{tikzpicture}

\definecolor{color0}{rgb}{1,0.498039215686275,0.0549019607843137}
\definecolor{color1}{rgb}{0.172549019607843,0.627450980392157,0.172549019607843}

\begin{axis}[
tick align=outside,
tick pos=left,
x grid style={white!69.0196078431373!black},
xmin=0.5, xmax=6.5,
xtick style={color=black},
xtick={1,2,3,4,5,6},
xticklabels={ITE\_best,\tmreg,\tmmed,\sdrreg,\sdrmed,Oracle},
y grid style={white!69.0196078431373!black},
ylabel={$\Delta$AUUC\_thout},
x tick label style={rotate=90,anchor=east},
ymin=173.750000000134, ymax=338.75000000003,
ytick style={color=black}
]
\addplot [black]
table {%
0.75 247.500000000087
1.25 247.500000000087
1.25 295.000000000077
0.75 295.000000000077
0.75 247.500000000087
};
\addplot [black]
table {%
1 247.500000000087
1 208.75000000005
};
\addplot [black]
table {%
1 295.000000000077
1 316.250000000003
};
\addplot [black]
table {%
0.875 208.75000000005
1.125 208.75000000005
};
\addplot [black]
table {%
0.875 316.250000000003
1.125 316.250000000003
};
\addplot [black, mark=*, mark size=1, mark options={solid,fill opacity=0}, only marks]
table {%
1 188.75000000002
1 181.250000000129
1 203.749999999884
1 316.250000000043
1 316.250000000064
1 321.250000000041
};
\addplot [black]
table {%
1.75 247.500000000087
2.25 247.500000000087
2.25 295.000000000077
1.75 295.000000000077
1.75 247.500000000087
};
\addplot [black]
table {%
2 247.500000000087
2 208.75000000005
};
\addplot [black]
table {%
2 295.000000000077
2 316.250000000003
};
\addplot [black]
table {%
1.875 208.75000000005
2.125 208.75000000005
};
\addplot [black]
table {%
1.875 316.250000000003
2.125 316.250000000003
};
\addplot [black, mark=*, mark size=1, mark options={solid,fill opacity=0}, only marks]
table {%
2 188.75000000002
2 181.250000000129
2 203.749999999884
2 316.250000000043
2 316.250000000064
2 321.250000000041
};
\addplot [black]
table {%
2.75 331.250000000034
3.25 331.250000000034
3.25 331.250000000034
2.75 331.250000000034
2.75 331.250000000034
};
\addplot [black]
table {%
3 331.250000000034
3 328.750000000036
};
\addplot [black]
table {%
3 331.250000000034
3 331.250000000034
};
\addplot [black]
table {%
2.875 328.750000000036
3.125 328.750000000036
};
\addplot [black]
table {%
2.875 331.250000000034
3.125 331.250000000034
};
\addplot [black, mark=*, mark size=1, mark options={solid,fill opacity=0}, only marks]
table {%
3 328.750000000036
3 326.250000000039
3 331.250000000035
3 331.250000000035
};
\addplot [black]
table {%
3.75 247.500000000087
4.25 247.500000000087
4.25 295.000000000077
3.75 295.000000000077
3.75 247.500000000087
};
\addplot [black]
table {%
4 247.500000000087
4 208.75000000005
};
\addplot [black]
table {%
4 295.000000000077
4 316.250000000003
};
\addplot [black]
table {%
3.875 208.75000000005
4.125 208.75000000005
};
\addplot [black]
table {%
3.875 316.250000000003
4.125 316.250000000003
};
\addplot [black, mark=*, mark size=1, mark options={solid,fill opacity=0}, only marks]
table {%
4 188.75000000002
4 181.250000000129
4 203.749999999884
4 316.250000000043
4 316.250000000064
4 321.250000000041
};
\addplot [black]
table {%
4.75 331.250000000034
5.25 331.250000000034
5.25 331.250000000034
4.75 331.250000000034
4.75 331.250000000034
};
\addplot [black]
table {%
5 331.250000000034
5 328.750000000036
};
\addplot [black]
table {%
5 331.250000000034
5 331.250000000034
};
\addplot [black]
table {%
4.875 328.750000000036
5.125 328.750000000036
};
\addplot [black]
table {%
4.875 331.250000000034
5.125 331.250000000034
};
\addplot [black, mark=*, mark size=1, mark options={solid,fill opacity=0}, only marks]
table {%
5 328.750000000036
5 326.250000000039
5 328.750000000036
5 331.250000000035
5 331.250000000035
};
\addplot [black]
table {%
5.75 331.250000000034
6.25 331.250000000034
6.25 331.250000000034
5.75 331.250000000034
5.75 331.250000000034
};
\addplot [black]
table {%
6 331.250000000034
6 331.250000000034
};
\addplot [black]
table {%
6 331.250000000034
6 331.250000000034
};
\addplot [black]
table {%
5.875 331.250000000034
6.125 331.250000000034
};
\addplot [black]
table {%
5.875 331.250000000034
6.125 331.250000000034
};
\addplot [black, mark=*, mark size=1, mark options={solid,fill opacity=0}, only marks]
table {%
6 331.250000000034
6 331.250000000033
6 331.250000000035
6 331.250000000035
};
\addplot [color0]
table {%
0.75 278.750000000066
1.25 278.750000000066
};
\addplot [color1, mark=triangle*, mark size=3, mark options={solid}, only marks]
table {%
1 270.906862745144
};
\addplot [color0]
table {%
1.75 278.750000000066
2.25 278.750000000066
};
\addplot [color1, mark=triangle*, mark size=3, mark options={solid}, only marks]
table {%
2 270.906862745144
};
\addplot [color0]
table {%
2.75 331.250000000034
3.25 331.250000000034
};
\addplot [color1, mark=triangle*, mark size=3, mark options={solid}, only marks]
table {%
3 330.71078431376
};
\addplot [color0]
table {%
3.75 278.750000000066
4.25 278.750000000066
};
\addplot [color1, mark=triangle*, mark size=3, mark options={solid}, only marks]
table {%
4 270.906862745144
};
\addplot [color0]
table {%
4.75 331.250000000034
5.25 331.250000000034
};
\addplot [color1, mark=triangle*, mark size=3, mark options={solid}, only marks]
table {%
5 330.661764705917
};
\addplot [color0]
table {%
5.75 331.250000000034
6.25 331.250000000034
};
\addplot [color1, mark=triangle*, mark size=3, mark options={solid}, only marks]
table {%
6 331.250000000034
};
\end{axis}

\end{tikzpicture}
\caption{$\Delta AUUC_{thout}$ (higher is better) of two \Teffect~models, corresponding \OurEstimator~models and Oracle model (theoretical truth). Box plots are done on 51 random splits, whiskers at 5/95 percentiles. Note how \OurEstimator~systematically increases the AUUC of standard \Teffect~estimators}
\label{fig:AUUC_thout}
\end{center}
\end{figure}
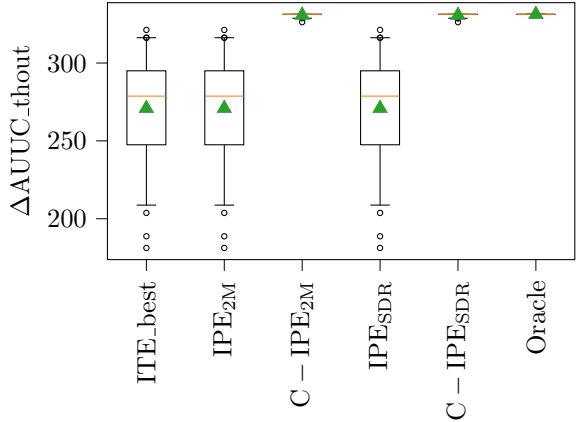

Figure \ref{fig:AUUC_thout} represents the performance of the same models as in Figure \ref{fig:AUUC} but with the $\mathrm{AUUC}_{\mathrm{thout}}$ metric (AUUC on a theoretical test set).

As expected, the Oracle has no variance because it does not $-$ by design $-$ suffer from the randomness of the training set and the $\mathrm{AUUC}_{\mathrm{thout}}$ metric gets rid of the variance of the test set.\\
\OurEstimator~models also have little to no $\mathrm{AUUC}_{\mathrm{thout}}$ variance in practice. This can be explained by the fact that these models learn treatment intake effect (\Meffect~) signal, which is far less noisy than the \Teffect~signal in low-compliance settings, and therefore suffer from less random variability.

\subsection{Hyper-parameters tuning}
For the open dataset CRITEO-UPLIFT1\footnote{\url{http://cail.criteo.com/criteo-uplift-prediction-dataset/}}, we first learn $\hat{\gamma}(x)$ $-$ modeling the true individual compliance $\compl(\user)= \Probdocgiv{\SCM}{\Treat}{\Display}{\user}$ probability $-$ from a separate process.
Namely, the $\hat{\gamma}(x)$ model is learned in a classical supervized learning way, as the compliance label $\Display$ is accessible. In terms of losses, we use a rebalanced LLH to account for the fact that classes $\Display=1$ and $\Display=0$ are very imbalanced. This step is crucial, as a good $\complE$ model ensures the superiority of \OurEstimator-based algorithms. The models are implemented using the logistic regression from Scikit-Learn \citep{pedregosa2011scikit}. The hyper-parameter grid considers (1) penalty regularization, (2) regularization strength and (3) the rebalancing (or not) of the LLH, as presented in Table~\ref{table:hyperparams1}.

\begin{table}[h]
\caption{Table of hyper-parameters for the compliance model}
\label{table:hyperparams1}
\centering
\begin{sc}
\begin{tabular}{c|c}
\toprule
parameter & values   \\                  \midrule
penalty regularization  & $\{L1, L2\}$ \\ 
$C$: inverse of regularization strength  &$\{0.01, 1, 10^2, 10^5\}$ \\          
$\text{class\_weight}$: weights associated with classes  &$\{\text{balanced}, \text{None}\}$  \\
\bottomrule
\end{tabular}
\end{sc}
\end{table}
Hyper-parameters are selected by internal 11-fold cross-validation on the training set. The best hyper-parameter triplet found is: L1 regularization, $C=100$ and $\text{class\_}\text{weight} = \text{'balanced'}$.

Then, we tune the two individual treatment effect models $\TEE{\Teffect}$ and $\TEE{\Meffect}$, by again performing internal cross-validation on the training set. We rank the hyper-parameters configuration according to the $\Delta$AUUC metric. The hyper-parameters grid considers (1) penalty regularization and (2) regularization strength, as presented in Table~\ref{table:hyperparams2}.
\begin{table}[h]
\caption{Table of hyper-parameters for the individual treatment effect models}
\label{table:hyperparams2}
\centering
\begin{sc}
\begin{tabular}{c|c}
\toprule
parameter & values   \\                  \midrule
penalty regularization  & $\{L1, L2\}$ \\ 
$C$: inverse of regularization strength  &$\{0.01, 0.1, 1, 10, 10^2, 10^3, 10^4, 10^5\}$ \\          
\bottomrule
\end{tabular}
\end{sc}
\end{table}

\paragraph{Remark}
For the plot of Figure \ref{fig:AUUC_real_dataset}, we only show the penalization L2, with an inverse regularization strength $C=10$ as we found that the $\Delta$AUUC was not sensitive to these parameters (see subscript $_\text{COMP}$ in Table~\ref{table:hyperparams_best}). For clarity, we display in Table~\ref{table:hyperparams_best} the best hyper-parameters tuples for each method, and the associated $\Delta$AUUC metric.

\begin{table}[h]
\caption{Table of hyper-parameters for the models. The metric is $\Delta$AUUC}
\label{table:hyperparams_best}
\centering
\begin{sc}
\begin{tabular}{l|ccc||ccc}
\toprule
method &	$C_{\text{best}}$ &	$\text{penalty}_{\text{best}}$ &	$\Delta\mathrm{AUUC}_{\text{best}}$ &	$C_{\text{comp}}$ &	$\text{penalty}_{\text{comp}}$ &	$\Delta\mathrm{AUUC}_{\text{comp}}$ \\
\midrule
\tmreg &	100000.0 &	L1 &	20484.044762 &	10.0 &	L2 &	20249.085940\\
\sdrreg &	1.0 &	L1 &	20596.476241 &	10.0 &	L2 &	20179.615970\\
\tmmed &	10.0 &	L2 &	21507.813960 &	10.0 &	L2 &	21507.813960\\
\sdrmed &	1.0 &	L2 &	21454.981632 &	10.0 &	L2 &	21333.441887\\
\bottomrule
\end{tabular}
\end{sc}
\end{table}

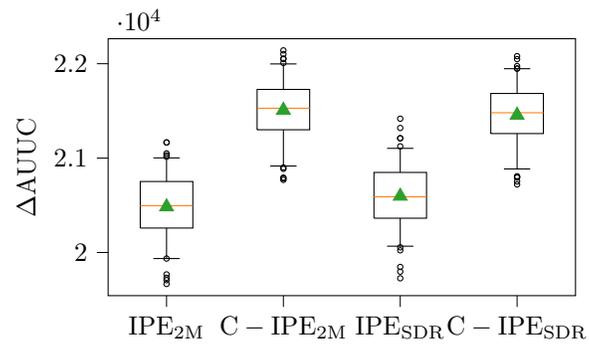
\begin{figure}[h]
\begin{center}
\pgfplotsset{width=.5\columnwidth,height=5cm}
\begin{tikzpicture}

\definecolor{color0}{rgb}{1,0.498039215686275,0.0549019607843137}
\definecolor{color1}{rgb}{0.172549019607843,0.627450980392157,0.172549019607843}

\begin{axis}[
tick align=outside,
tick pos=left,
x grid style={white!69.0196078431373!black},
xmin=0.5, xmax=4.5,
xtick style={color=black},
xtick={1,2,3,4},
xticklabels={\tmreg,\tmmed,\sdrreg,\sdrmed},
y grid style={white!69.0196078431373!black},
ylabel={$\Delta$AUUC},
ymin=19542.5772368098, ymax=22264.968889994,
ytick style={color=black}
]
\addplot [black]
table {%
0.775 20258.9488883814
1.225 20258.9488883814
1.225 20752.0818751649
0.775 20752.0818751649
0.775 20258.9488883814
};
\addplot [black]
table {%
1 20258.9488883814
1 19935.9017952068
};
\addplot [black]
table {%
1 20752.0818751649
1 21002.2286227214
};
\addplot [black]
table {%
0.8875 19935.9017952068
1.1125 19935.9017952068
};
\addplot [black]
table {%
0.8875 21002.2286227214
1.1125 21002.2286227214
};
\addplot [black, mark=*, mark size=1, mark options={solid,fill opacity=0}, only marks]
table {%
1 19731.9247669926
1 19767.5656656416
1 19935.8329663509
1 19666.3223119546
1 19706.7208370387
1 21165.9887265199
1 21050.3536219712
1 21014.475096467
1 21165.8921487409
1 21030.7585790319
};
\addplot [black]
table {%
1.775 21300.2973696449
2.225 21300.2973696449
2.225 21727.7923795837
1.775 21727.7923795837
1.775 21300.2973696449
};
\addplot [black]
table {%
2 21300.2973696449
2 20916.3718533299
};
\addplot [black]
table {%
2 21727.7923795837
2 21997.8314412338
};
\addplot [black]
table {%
1.8875 20916.3718533299
2.1125 20916.3718533299
};
\addplot [black]
table {%
1.8875 21997.8314412338
2.1125 21997.8314412338
};
\addplot [black, mark=*, mark size=1, mark options={solid,fill opacity=0}, only marks]
table {%
2 20793.4691173767
2 20886.2792998885
2 20900.5419100536
2 20769.476556348
2 20784.4315193847
2 22054.7799871318
2 22049.2366974415
2 22008.0094826856
2 22141.2238148492
2 22102.1235537939
};
\addplot [black]
table {%
2.775 20363.3252518927
3.225 20363.3252518927
3.225 20848.2786048579
2.775 20848.2786048579
2.775 20363.3252518927
};
\addplot [black]
table {%
3 20363.3252518927
3 20066.7772978749
};
\addplot [black]
table {%
3 20848.2786048579
3 21103.3669618649
};
\addplot [black]
table {%
2.8875 20066.7772978749
3.1125 20066.7772978749
};
\addplot [black]
table {%
2.8875 21103.3669618649
3.1125 21103.3669618649
};
\addplot [black, mark=*, mark size=1, mark options={solid,fill opacity=0}, only marks]
table {%
3 19846.5551967057
3 20022.8999060532
3 20054.6420798915
3 19796.8612293238
3 19726.7077341344
3 21417.6502940861
3 21210.3015738564
3 21124.6560350396
3 21210.6772180513
3 21319.2761480061
};
\addplot [black]
table {%
3.775 21259.9792521914
4.225 21259.9792521914
4.225 21684.9398638088
3.775 21684.9398638088
3.775 21259.9792521914
};
\addplot [black]
table {%
4 21259.9792521914
4 20885.0572947264
};
\addplot [black]
table {%
4 21684.9398638088
4 21947.2096848439
};
\addplot [black]
table {%
3.8875 20885.0572947264
4.1125 20885.0572947264
};
\addplot [black]
table {%
3.8875 21947.2096848439
4.1125 21947.2096848439
};
\addplot [black, mark=*, mark size=1, mark options={solid,fill opacity=0}, only marks]
table {%
4 20793.9602368941
4 20800.0483623202
4 20807.8351256496
4 20720.9414171959
4 20753.2061823692
4 21975.5685920296
4 21950.0693779948
4 22080.1793254081
4 22048.3937753226
4 21947.7273749945
};
\addplot [color0]
table {%
0.775 20495.5834048916
1.225 20495.5834048916
};
\addplot [color1, mark=triangle*, mark size=3, mark options={solid}, only marks]
table {%
1 20484.0447616421
};
\addplot [color0]
table {%
1.775 21528.3438582824
2.225 21528.3438582824
};
\addplot [color1, mark=triangle*, mark size=3, mark options={solid}, only marks]
table {%
2 21507.8139601049
};
\addplot [color0]
table {%
2.775 20589.9548954387
3.225 20589.9548954387
};
\addplot [color1, mark=triangle*, mark size=3, mark options={solid}, only marks]
table {%
3 20596.4762409374
};
\addplot [color0]
table {%
3.775 21480.4763805099
4.225 21480.4763805099
};
\addplot [color1, mark=triangle*, mark size=3, mark options={solid}, only marks]
table {%
4 21454.981631828
};
\end{axis}

\end{tikzpicture}
\caption{$\Delta AUUC$ (higher is better) for the best hyper-parameter configurations. Box plots are done on 100 random splits, whiskers at 5/95 percentiles. Note how \OurEstimator~systematically increases the $\Delta$AUUC of standard \Teffect~estimators}
\label{fig:AUUC_best}
\end{center}
\end{figure}

The corresponding performance the the best configuration are presented in Figure~\ref{fig:AUUC_best}.

Note that the order of magnitude for each method is very similar with those presented in Figure~\ref{fig:AUUC_real_dataset}, \ie~for penalization L2 and $C=10$, which shows that the fine tuning of the hyper-parameters has a negligible impact on graphical displays.

\end{document}